\documentclass[lettersize,journal]{IEEEtran}
\usepackage{xcolor}  
\definecolor{myPurple}{RGB}{128,0,128}  

\usepackage{amsfonts}
\usepackage{bm}
\usepackage{amsmath}
\usepackage{graphicx}
\usepackage{subcaption}
\usepackage[normalem]{ulem}
\usepackage{multirow}
\usepackage{amsmath,amsfonts}
\usepackage{algorithmic}
\usepackage{algorithm}
\usepackage{array}
\usepackage{textcomp}
\usepackage{stfloats}
\usepackage{url}
\usepackage{verbatim}
\usepackage{cite}
\usepackage{tabularx}  
\usepackage{flushend}
\usepackage{xcolor}
\usepackage{url}
\usepackage{colortbl}
\usepackage{adjustbox} 
\usepackage{soul}

\definecolor{orange}{HTML}{FCE5D6}

\usepackage[labelformat=simple]{subcaption}

\makeatletter
\def\testclr#1#{\@testclr{#1}}
\def\@testclr#1#2{{\fboxsep\z@\fbox{\colorbox#1{#2}{\phantom{XX}}}}}

\makeatother

\begin{document}

\title{CURL-SLAM: \\ Continuous and Compact LiDAR Mapping}

\author{Kaicheng Zhang$^{1}$, Shida Xu$^{1}$, Yining Ding$^2$, Xianwen Kong$^{2}$ and Sen Wang$^{1}$
\thanks{$^{1}$ Department of Electrical and Electronic Engineering, Imperial College London, W12 0BZ, UK\\
{\tt\small k.zhang23@alumni.imperial.ac.uk}\\
{\tt\small \{s.xu23, sen.wang\}@imperial.ac.uk}}%
\thanks{$^{2}$School of Engineering and Physical Sciences, Heriot-Watt University, Edinburgh, EH14 4AS, UK
		{\tt\small \{yd2007, x.kong\}@hw.ac.uk}}%
\thanks{Corresponding author: Sen Wang}
}



\maketitle

\begin{abstract}
	This paper studies 3D LiDAR mapping with a focus on developing an updatable and localizable map representation that enables continuity, compactness and consistency in 3D maps.
	Traditional LiDAR Simultaneous Localization and Mapping (SLAM) systems often rely on 3D point cloud maps, which typically require extensive storage to preserve structural details in large-scale environments.
	In this paper, we propose a novel paradigm for LiDAR SLAM by leveraging the Continuous and Ultra-compact Representation of LiDAR (CURL) introduced in \cite{Zhang-RSS-22}. 
	Our proposed LiDAR mapping approach, CURL-SLAM, produces compact 3D maps capable of continuous reconstruction at variable densities using CURL's spherical harmonics implicit encoding, and achieves global map consistency after loop closure.
	Unlike popular Iterative Closest Point (ICP)-based LiDAR odometry techniques, CURL-SLAM formulates LiDAR pose estimation as a unique optimization problem tailored for CURL and extends it to local Bundle Adjustment (BA), enabling simultaneous pose refinement and map correction.
	Experimental results demonstrate that CURL-SLAM achieves state-of-the-art 3D mapping quality and competitive LiDAR trajectory accuracy, delivering sensor-rate real-time performance (10 Hz) on a CPU. We will release the CURL-SLAM implementation to the community.

\end{abstract}


\begin{figure}[t]
	\includegraphics[width=\linewidth]{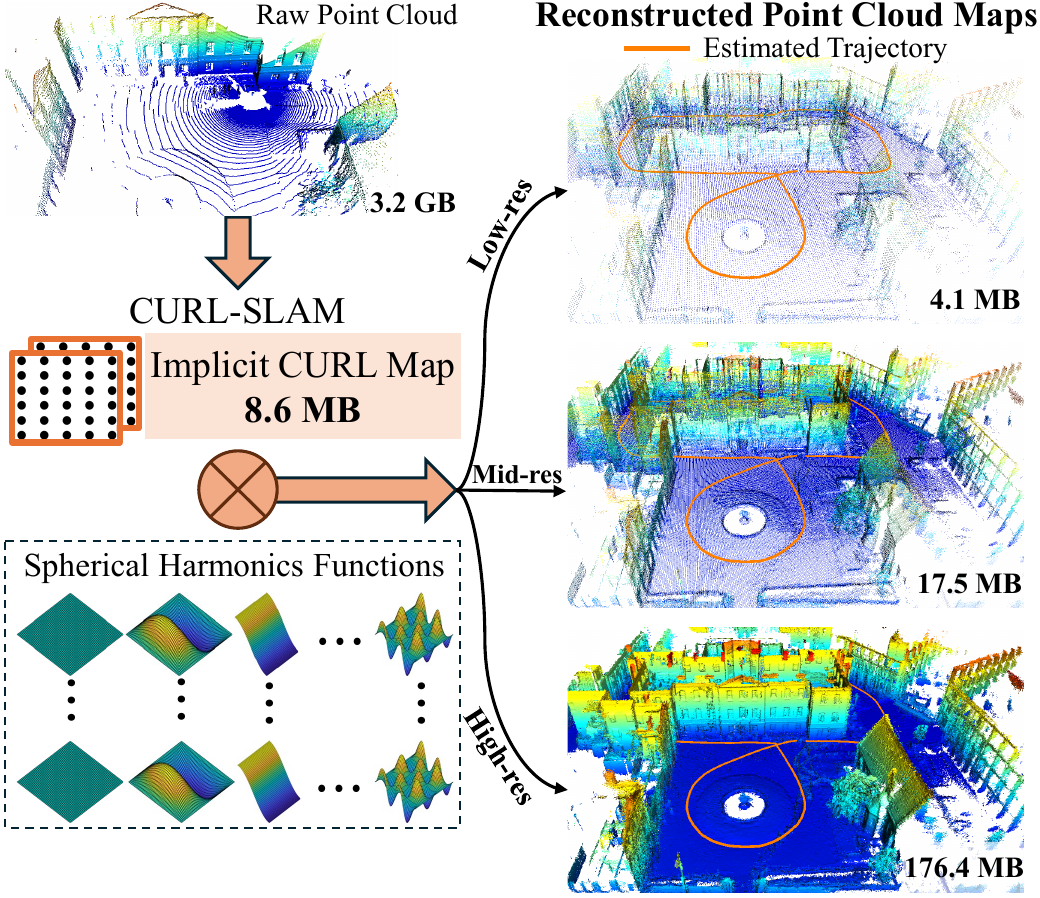}
	\caption{Globally consistent 3D maps reconstructed by CURL-SLAM. Point cloud maps with different resolutions are continuously reconstructed using the same CURL map which is ultra-compact ($0.26\%$ of the $3.2$ GB raw point clouds). 
	}
	\label{fig:cover}
	\vspace{-6pt}
\end{figure}



\section{Introduction}

3D mapping is pivotal for Simultaneous Localization and Mapping (SLAM) and autonomous robots. The design and performance of 3D mapping algorithms are significantly affected by their map representations. Among the various 3D map representations, point clouds stand out as the most prevalent due to their simplicity and extensive algorithms and tools available for processing, such as the Point Cloud Library (PCL). However, most point cloud based odometry/SLAM systems, e.g., LOAM \cite{zhang2014loam} and its variants \cite{shan2018lego,shan2020lio}, downsample their point clouds for mapping due to computational and storage efficiency, resulting in a potential loss of map resolution and 3D structural details. At the same time, storing all points is not ideal, as map size grows exponentially, leading to significant processing overheads and ample storage space.

Other map representations, such as surfels and meshes, are popular in several SLAM systems \cite{whelan2015elasticfusion,vizzo2021poisson}. {Surfel-based representations benefit SLAM by enabling efficient incremental updates and correcting local distortions.} However, the level of detail in surfel-based maps heavily depends on the number of surfels used, {which makes large-scale mapping more challenging}.
{Mesh representations, widely used in 3D modeling, offer notable memory efficiency for 3D mapping, especially in structured environments. They also enhance SLAM systems by supporting dense surface reconstruction \cite{vizzo2021poisson,ruan2023slamesh,lin2023immesh}. However, achieving high accuracy in mesh maps often requires introducing a large number of vertices. Furthermore, updating a mesh map can be computationally expensive for a real-time SLAM system, as it involves re-establishing vertex connectivity.}


Recently, various neural implicit map representations have emerged, {typically modeling the spatial environment as a continuous function over coordinates.} These methods enable continuous reconstruction, occupancy prediction \cite{mescheder2019occupancy}, and shape completion using Signed Distance Functions (SDFs) \cite{park2019deepsdf}. {Such implicit representations offer smoother, scalable, and memory-efficient reconstructions compared to traditional point-cloud maps. These advantages provide the potential for more accurate, globally consistent, and easily updatable SLAM systems.}
Several LiDAR-based SLAM systems have adopted neural implicit representations, such as NeRF-LOAM \cite{deng2023nerf}, LONER \cite{isaacson2023loner}, and PIN-SLAM \cite{pan2024pin}. Notably, PIN-SLAM achieves state-of-the-art LiDAR odometry performance and is among the few complete LiDAR SLAM systems capable of generating globally consistent maps in real-time. {A globally consistent map effectively eliminates ghosting artifacts, particularly in revisited regions within large-scale environments.} However, these neural implicit approaches generally require substantial computational resources, typically GPU acceleration, limiting their practicality for many robotic applications.


The CURL representation \cite{Zhang-RSS-22} was recently introduced as a method to implicitly encode 3D LiDAR points using spherical harmonics functions. This approach provides a highly compact map representation while enabling continuous reconstruction. Its extension, CURL-MAP \cite{zhang2024curl}, further explored the potential of the CURL representation for odometry and mapping. Despite these developments, CURL-MAP is still limited in its ability to achieve consistent large-scale mapping, due to its lack of loop closure and local-window optimization, which are necessary for handling drift in large environments.

In this paper, we present CURL-SLAM, the first full LiDAR SLAM system based on the CURL representation, capable of generating an ultra-compact, updatable, and globally consistent 3D map.
CURL-SLAM enables real-time pose estimation and continuous 3D mapping on a CPU at a 10 Hz LiDAR frame rate. Our main contributions are as follows:

\begin{itemize}
	\item A novel LiDAR SLAM system that allows continuous 3D mapping with adjustable reconstruction resolutions from a single compact CURL map. CURL-SLAM formulates pose estimation as a unique optimization problem specifically tailored for CURL, distinguishing it from widely used Iterative Closest Point (ICP) based techniques.
	\item A rigorous formulation of local Bundle Adjustment (BA) that jointly optimizes both the CURL map and poses, enhancing map consistency and accuracy. To facilitate effective BA optimization, we derive the analytical Jacobian matrices in detail.
	\item Extensive experimental evaluations demonstrating CURL-SLAM's state-of-the-art mapping performance, particularly in terms of map accuracy and compactness.
\end{itemize}

As the example shown in Fig. \ref{fig:cover}, CURL-SLAM's unique use of the CURL representation enables the recovery of accurate 3D maps at various resolutions from a single CURL implicit map, whose size is only $0.26\%$ of the raw point clouds.
We will release our code as open-source for the community\footnote{\url{https://github.com/SenseRoboticsLab/CURL-SLAM}}.

\noindent\textbf{Extension to CURL-MAP \cite{zhang2024curl}}: The proposed CURL-SLAM addresses the limitations of CURL-MAP \cite{zhang2024curl}, with a specific focus on achieving globally consistent maps. To this end, this paper introduces novel CURL-based local BA and loop closure mechanisms, representing significant extensions to CURL-MAP \cite{zhang2024curl}. Furthermore, we propose new methods for patch generation, data association and height-augmented image formation, replacing the quasi-conformal parametrization used in CURL-MAP \cite{zhang2024curl}. These advancements dramatically reduce the computational overhead of constructing and updating CURL-based maps in large-scale environments, allowing for efficient map creation and real-time updates on CPU-based systems. Additionally, the experiments have been completely redesigned, featuring evaluations against three new state-of-the-art algorithms to provide robust benchmarking.




\section{Related Work}

\subsection{3D Map Representation}

{Traditional point cloud maps inherently suffer from high memory consumption, susceptibility to sensor noise, and difficulty in maintaining consistency over large-scale environments. Non-point cloud representations, including continuous implicit maps and neural-based representations, offer smoother, scalable, and memory-efficient reconstructions. These properties enable more accurate, globally consistent, and easily updatable SLAM systems. However, neural-based implicit representations typically require significant computational resources, such as GPU acceleration, which can limit their practicality for real-time applications.}

\subsubsection{Continuous Representation}
SDF is widely utilized as continuous implicit map representations in robotics and computer vision due to their ability to encode spatial information efficiently. SDF can be categorized into two main types: Truncated Signed Distance Function (TSDF) and Euclidean Signed Distance Function (ESDF)  \cite{oleynikova2016signed}.

TSDF maps are commonly employed in range mapping, especially for RGB-D and LiDAR sensors. To improve mapping efficiency, the weighted distance from each voxel to the nearest surface along the sensor ray is used to update the TSDF map \cite{millane2018c,newcombe2011kinectfusion,newcombe2015dynamicfusion,vespa2018efficient}. On the other hand, ESDF maps, which {store} the Euclidean distance to its nearest surface for each voxel, are preferred in path planning applications \cite{zhou2019robust}.
To generate ESDF maps, several works\cite{reijgwart2019voxgraph,millane2021freetures,oleynikova2017voxblox}  have proposed methods to convert from TSDF representations.
Most TSDF implementations require GPU acceleration and suffer from memory efficiency limitations. VDBFusion \cite{vizzo2022vdbfusion} has been introduced to make it {run} on a CPU more efficiently. However, due to the nature of SDF maps being stored as discretized values within voxel grids, their accuracy is highly dependent on the grid resolution, which can lead to significant memory requirements for dense detailed mapping.
	{Nevertheless, SDF-based representations offer notable benefits for SLAM systems, including incremental and continuous mapping capabilities, high accuracy in surface reconstruction, and efficient integration of sensor measurements over time.}

With neural networks, the function of SDF maps can be continuously represented, eliminating the need for voxel grids and offering a more compact and scalable representation \cite{park2019deepsdf, zhong2023shine}. Occupancy Networks \cite{mescheder2019occupancy} offer a probabilistic approach to querying occupancy at any point in Euclidean space, enhancing the flexibility of implicit representations. However, extracting the surface implicitly encoded by neural networks requires initializing a voxel grid or using a ray-casting method to produce a signed distance field or an occupied grid.
	{Recent advancements like Neural Radiance Fields (NeRF) \cite{mildenhall2021nerf} and 3D Gaussian Splatting (3DGS) \cite{kerbl20233d} enhance implicit map representations with rendering new view points.} However, these methods need to be executed on GPUs and face challenges in accurately representing 3D geometry in large-scale environments.

	{Kernel-based non-parametric} point cloud modelling techniques, such as Gaussian Processes \cite{o2012gaussian} and Hilbert Space methods \cite{ramos2016hilbert}, have also been explored. However, these approaches typically demand high computational resources, limiting their practicality for real-time applications.
Gaussian Mixture Models (GMMs) have also been applied to continuous 3D reconstruction and are effective in generating compact maps \cite{li2024gmmap}, capable of running on low-power CPUs to produce continuous occupancy representations. However, they do not focus on highly detailed reconstruction. In contrast, \cite{goel2024gira} and \cite{goel2023probabilistic} employ 4D Gaussians to encode point clouds with intensity, achieving compact storage and detailed reconstruction, but at the cost of longer GPU training times, which poses challenges for real-time use. 

	{FS-SLAM \cite{zhao20212d} inspired our approach by utilizing Fourier Series functions to encode the map, achieving continuous and compact reconstructions. However, FS-SLAM is limited to 2D representations and pillar-like environments, significantly restricting its practical applicability to general 3D mapping tasks. In contrast, our method extends this concept to general 3D environments by leveraging spherical harmonics as the underlying functional representation. This enables continuous, detailed, and computationally efficient reconstructions suitable for broader real-time SLAM applications.}

	{Specifically, CURL-SLAM integrates spherical harmonics with a spatial hashing grid, efficiently clustering point clouds into manageable patches. This reduces computational complexity and memory consumption compared to neural implicit and GMM-based representations. Unlike many neural or occupancy-based methods, our method operates efficiently on CPUs without requiring GPU acceleration, thus balancing computational efficiency and reconstruction fidelity at arbitrary resolutions.}

\subsubsection{Compact Representation}

For traditional compact representations, occupancy maps, e.g., OctoMap \cite{hornung2013octomap} and Supereight \cite{vespa2018efficient}, are widely adopted due to their compactness and memory efficiency by employing octrees for spatial partitioning. Wavemap \cite{reijgwart2023efficient} further compresses maps by storing wavelet coefficients within the octrees. Recently, GMM-based occupancy map representations \cite{li2024gmmap, goel2024gira, goel2023probabilistic} also achieve compact memory footprint by only storing Gaussian parameters along with their weights.

For learning-based methods that cooperate with the aforementioned occupancy and SDF maps, memory usage can be further reduced by relying solely on network parameters to represent 3D geometry. For occupancy maps, Occupancy Networks \cite{mescheder2019occupancy} utilize neural networks to implicitly encode the occupancy probability function independently of voxel size, thereby achieving compact representations. However, occupancy maps face challenges in recovering structural details in large-scale scenarios, as the grid size inherently limits their resolution. Similarly, SDF maps encoded by neural networks \cite{park2019deepsdf,sucar2021imap,ortiz2022isdf,zhu2022nice,sandstrom2023point} can produce smoother surfaces, but they also encounter similar issues since the surface needs to be extracted from voxelized signed distance fields, which limits detail fidelity.

	{Compact representations significantly benefit SLAM systems by enabling efficient memory usage, facilitating scalable mapping of extensive environments, and supporting rapid map updates crucial for real-time operations. However, many existing compact methods, particularly those using neural networks, require GPU acceleration or exhibit limited reconstruction accuracy. In contrast, CURL-SLAM combines compactness and efficiency by encoding surfaces using spherical harmonics and organizing spatial information through a hashing grid. This approach provides detailed and scalable reconstructions while maintaining computational efficiency suitable for real-time SLAM on CPUs.}


\subsection{LiDAR Odometry and SLAM}

In prevailing LiDAR odometry and mapping frameworks, such as LOAM \cite{zhang2014loam}, LeGO-LOAM \cite{shan2018lego}, LIO-SAM \cite{shan2020lio}, CT-ICP \cite{deschaud2021ct}, KISS-ICP \cite{vizzo2023kiss}, and FAST-LIO2 \cite{xu2022fast}, the focus has largely been on using pose optimization to enhance the accuracy of point cloud maps. These methods achieve high localization accuracy and robustness in real-time performance. {Furthermore, BALM \cite{liu2021balm} and BALM 2.0 \cite{liu2023efficient} achieve locally consistent maps on small-scale environments, while HBA \cite{liu2023large} was proposed for achieving globally consistent maps. All these methods rely on further trajectory optimization to improve consistency.} Although point cloud maps are simple and straightforward, they inherently contain sensor noise, consume large amounts of memory, and are challenging to update. 

	{Non-point cloud map-based SLAM frameworks have been proposed to address these limitations. For instance,} ElasticFusion \cite{whelan2015elasticfusion}, to some extent, paved the way for vision-based map-centric methodologies by maintaining a deformation node list on a map instead of a pose graph. This idea has been extended to LiDAR \cite{park2021elasticity, park2018elastic}, {this surfel-based representation benefits SLAM by facilitating incremental updates and correcting local distortions through the deformation graph. However, the increasing computational complexity with deformation nodes makes detailed mapping challenging for large-scale environments.}

Recently, several LiDAR SLAM systems employing simplified representations have been proposed, including SegMap \cite{dube2020segmap}, Quadric Representation \cite{xia2023quadric}, VoxelMap \cite{yuan2022efficient}, VoxelMap++ \cite{wu2023voxelmap++}, and SLIM \cite{yu2024slim}. {These frameworks use simplified geometric features for map representation, significantly benefiting SLAM systems by reducing storage requirements and computational complexity, thus enabling faster pose estimation. However, this simplification primarily facilitates efficient localization rather than accurate and detailed 3D reconstruction.}

Mesh based SLAM systems, {such as PUMA \cite{vizzo2021poisson}, SLAMesh \cite{ruan2023slamesh}, and ImMesh \cite{lin2023immesh}, explicitly encode surface geometry into mesh structures. These methods benefit SLAM systems by providing dense, detailed geometric reconstructions beneficial for robust localization and precise scene analysis. However, real-time performance is limited in PUMA due to computational demands, and while SLAMesh and ImMesh achieve real-time operations, they do not incorporate loop closure. Furthermore, handling loop closure in mesh-based systems typically involves complicated procedures for disconnecting and re-establishing vertex connections.}

LiDAR SLAM systems employing implicit map representations have garnered attention in recent years. IMLS-SLAM \cite{deschaud2018imls} utilizes implicit moving least squares for map representation, {beneficial for accurate map-based localization}, but faces high computational demands.  {Incremental neural implicit SLAM frameworks, such as} LONER \cite{isaacson2023loner} and NeRF-LOAM \cite{deng2023nerf}, {leverage neural representations to facilitate high-fidelity reconstruction and robust pose estimation}, though only LONER achieves real-time performance. Our previous work, CURL-MAP \cite{zhang2024curl}, encodes an implicit map using spherical harmonics functions, allowing for continuous reconstruction and real-time odometry. However, none of these methods incorporate loop closure, preventing the creation of globally consistent maps. PIN-SLAM \cite{pan2024pin} bridges this gap by using neural points to compactly encode SDF, achieving state-of-the-art performance with the ability to build a globally consistent map while performing real-time odometry on GPUs.

Despite these advancements, building a SLAM system that is capable of building globally consistent, compact and continuous maps in real-time using CPUs remains a considerable challenge. {The proposed CURL-SLAM addresses these limitations by providing detailed implicit reconstructions using spherical harmonics functions combined with spatial hashing grids, achieving computational efficiency and global consistency without the need for GPU resources.}

\section{CURL-based Odometry and Mapping}
\label{sec:approach overview}
\begin{figure*}
	\includegraphics[width=\linewidth]{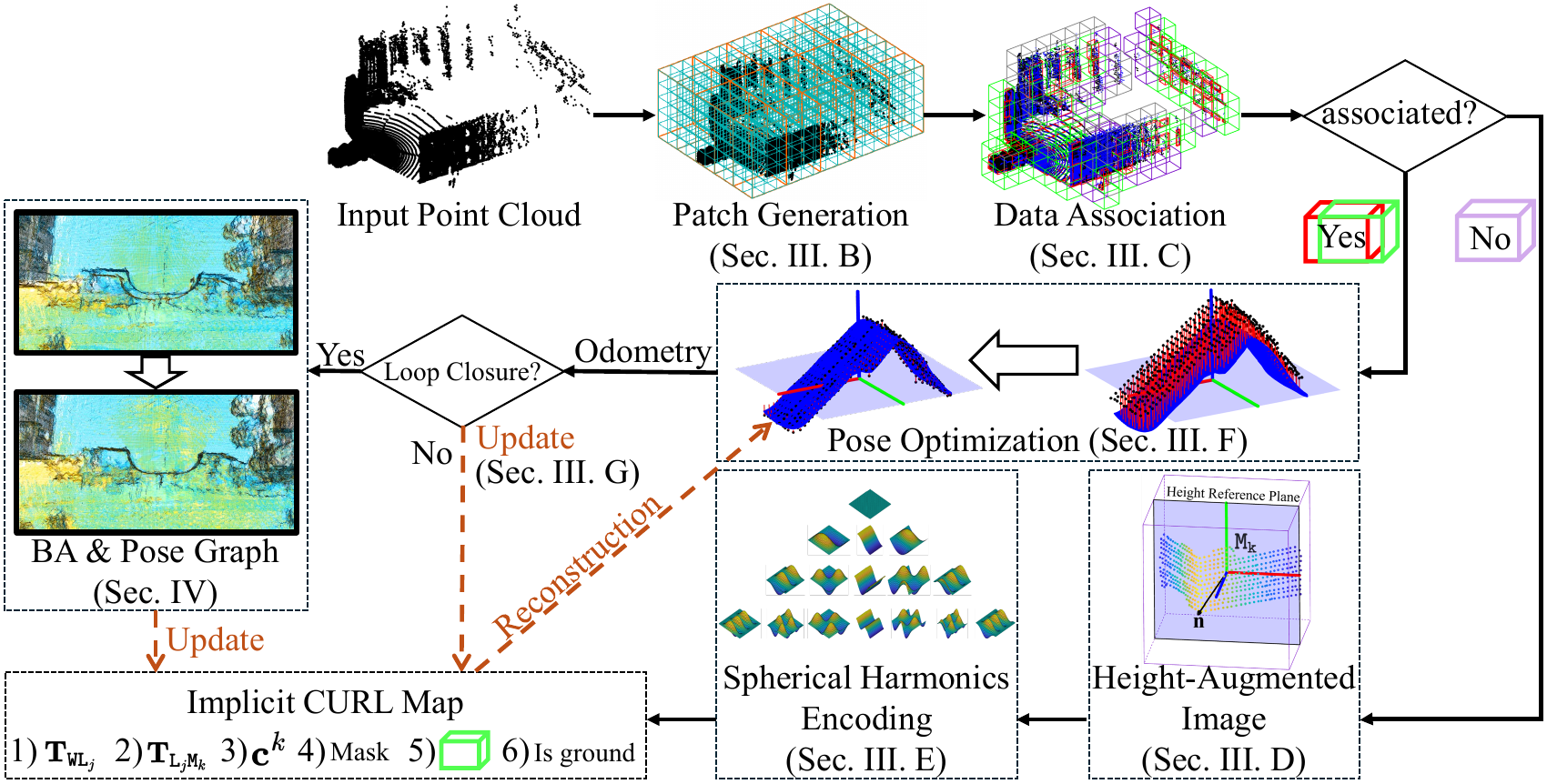}
	\caption{Pipeline of the proposed CURL-SLAM method.}
	\label{fig:framework}
\end{figure*}

{The full pipeline of our CURL-SLAM method is shown in Fig. \ref{fig:framework}. The SLAM system comprises four main modules: data association, pose optimization, map encoding, and local BA with pose graph optimization, which together enable the construction of a globally consistent map}.
Instead of using the popular planar/edge features detected from LiDAR point clouds or ICP methods, like LOAM \cite{zhang2014loam} and its variants, our method implicitly encodes point clouds as spherical harmonics functions, using the CURL representation proposed in \cite{Zhang-RSS-22}, for odometry and mapping. In this section, we describe the odometry and mapping modules of CURL-SLAM.


\subsection{Notation}

A typeface $\mathbb{A}$ denotes a set, bold capital letters $\mathbf{A}$ represent matrices, bold lowercase letters $\mathbf{a}$ denote vectors, and unbolded letters $a$ represent scalars.
$\mathbf{T}_{\mathtt{A}\mathtt{B}}\in \mathbb{R}^{4 \times 4}$ represents the transformation from the coordinate frame $\mathtt{B}$ to the coordinate frame $\mathtt{A}$, while ${}_\mathtt{A}\mathbf{p}_{i}\in \mathbb{R}^{3\times 1}$ refers to the $i$-th point in the coordinate frame $\mathtt{A}$.
The transformation of a point from frame $\mathtt{A}$ to frame $\mathtt{B}$ is defined as ${}_{\mathtt{B}}\mathbf{p}_i = \mathbf{T}_{\mathtt{BA}}\ {}_{\mathtt{A}}\mathbf{p}_i$, omitting the need for explicit homogeneous coordinates for simplicity.

\subsection{Patch Generation}
\label{sec:patch generation}

Since ground and non-ground points from LiDAR scans provide different constraints to pose estimation and vary in their complexity for mapping, the current LiDAR scan $j$ is pre-segmented into ground and non-ground points. We use patchwork++ \cite{lee2022patchworkpp} in this work due to its efficiency. Meanwhile, the scan is segmented into a set of {raw scan} patches $\{M^k \} \subseteq \mathbb{M}_j$ along with their corresponding points $\{{}_{\mathtt{L}_j}\mathbf{P}^k \in \mathbb{R}^{3 \times \psi} \}$ in $j$-th LiDAR frame $\mathtt{L}_j$, using a simple yet efficient volumetric clustering method, i.e., the LiDAR's local 3D space is pre-divided into fixed-size cubic voxels and the set of LiDAR points that fall within the same voxel is clustered as a patch, as the blue cubes shown in Fig. \ref{fig:patch generation}.



\subsection{Data Association of Patches}
\label{sec:data association}

\begin{figure}
	\centering
	\begin{subfigure}{0.50\linewidth}
		\includegraphics[width=1\linewidth]{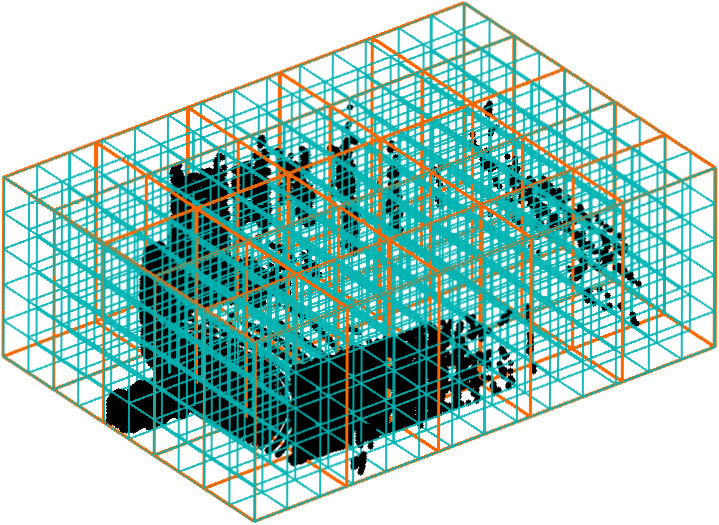}
		\caption{Patch generation.}
		\label{fig:patch generation}
	\end{subfigure}
	\begin{subfigure}{0.48\linewidth}
		\includegraphics[width=1\linewidth]{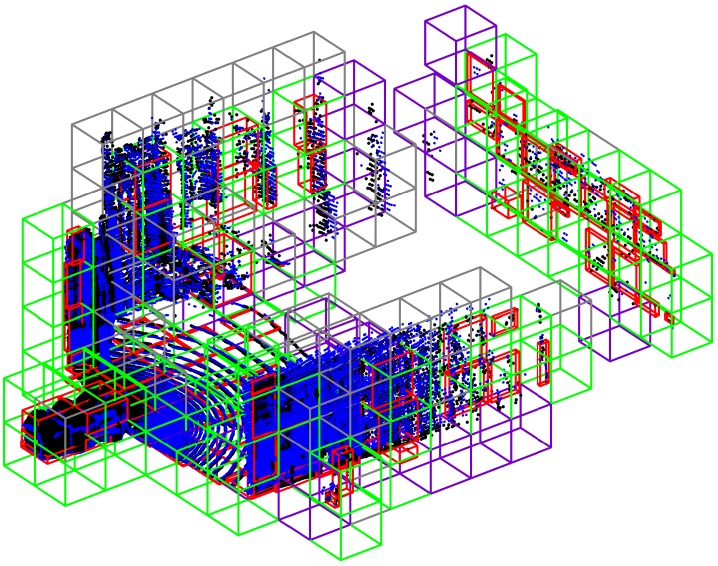}
		\caption{Patch association.}
		\label{fig:data association}
	\end{subfigure}
	\caption{Patch generation and association;
		(a) Blue and orange grids show the cubic voxels and the regions of a scan;
		(b) Selected successful associations between  {raw} scan patches (red boxes) and submap patches (green boxes) for pose estimation and {submap} {updates}. Purple boxes represent new patches to be inserted into the {submap}. 
	}

	\label{fig:data asso}
\end{figure}

For each patch $M^k$, it is necessary to determine whether it is associated with a patch in the current submap {$\mathbb{S}$} {as odometry is computed solely based on the current submap. The details of how this submap is constructed will be introduced later}.
Assume $\mathbf{T}_{\mathtt{W}\mathtt{L}_j}$, the transformation from frame $\mathtt{L}_j$ to the world frame $\mathtt{W}$ is known, the patch points ${}_{\mathtt{L}_j}\mathbf{P}^k$ can be transformed into $\mathtt{W}$ via $\mathbf{T}_{\mathtt{W}\mathtt{L}_j}$.
Once being transformed into $\mathtt{W}$, the world frame aligned bounding box of $M^k$ is then computed by finding its minimum and maximum values along the Euclidean axes of $\mathtt{W}$ shown as the red boxes in Fig. \ref{fig:data association}.
Intersection over Union (IoU) is employed to identify potential associations from $M^k$'s nearby patches in {$\mathbb{S}$}. 
If $M^k$ has multiple inlier associations, the associated {submap} patch{es} in {$\mathbb{S}$} with the highest IoU score is selected.

To reduce the computational complexity of the subsequent modules, $\mathbb{M}_j$ of a scan is spatially refined within regions, illustrated as the orange grids in Fig. \ref{fig:patch generation}. Only the top $\beta_{ng}$ and $\beta_{g}$ associations which have the highest IoU scores for non-ground and ground patches, respectively, are selected from each region for both pose estimation (Section \ref{sec:Pose Optimisation}) and {submap} update (Section \ref{sec: update_sph}), shown as the red and green boxes in Fig. \ref{fig:data association}{, and rest of the associations in black boxes are used for {submap} update only.}

Purple boxes in Fig. \ref{fig:data association} are those $M^k$ without associations from the submap going to be inserted in the {submap} after generating their CURL representation which has the following attributes:
$1)$ $\mathbf{T}_{\mathtt{W}\mathtt{L}_j}$;
$2)$ $\mathbf{T}_{\mathtt{L}_j\mathtt{M}_k}$;
$3)$ spherical harmonics coefficients $\mathbf{c}^k$ of $M^k$ (Section \ref{sec:spherical harmonics function encoding});
$4)$ mask;
$5)$ axis-aligned bounding box in $\mathtt{W}$;
$6)$ ground or non-ground label.
$\mathbf{T}_{\mathtt{L}_j\mathtt{M}_k}$ is the transformation from $M^k$'s patch frame $\mathtt{M}_k$ to the local LiDAR frame.
	{Then} we describe how to compute the 2) and 3) attributes before formulating the CURL-based pose optimization for 1).

\begin{figure}[t]
	\centering
	\begin{subfigure}{0.54\linewidth}
		\includegraphics[width=1\linewidth]{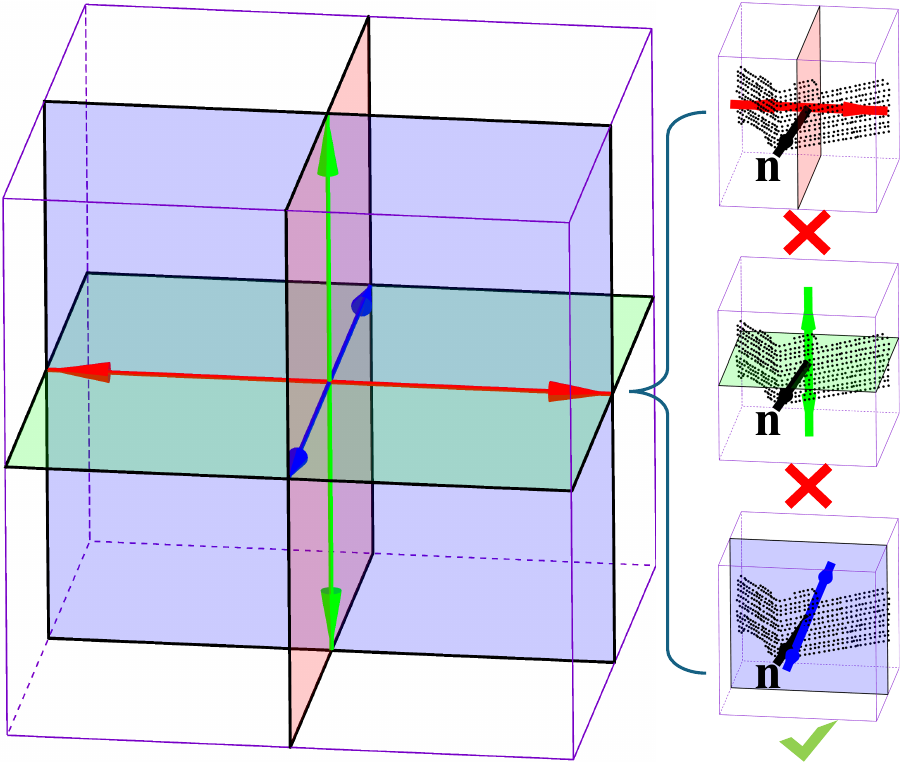}
		\caption{Projection plane selection}
		\label{fig:projection_planes}
	\end{subfigure}
	\begin{subfigure}{0.44\linewidth}
		\includegraphics[width=1\linewidth]{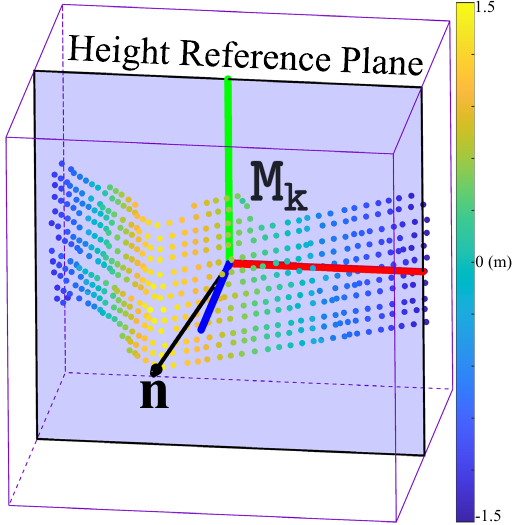}
		\caption{Patch frame $\mathtt{M}_k$}
		\label{fig:patch_coordinate}
	\end{subfigure}
	\caption{(a) Three axis-aligned orthogonal reference planes and their normals in a voxel. $\mathbf{n}$ is the normal vector of the patch points which used to select the height reference plane;
		(b) Patch frame $\mathtt{M}_k$ defined on the height reference plane.
	}
\end{figure}

\subsection{Height-Augmented Image}
\label{sec:Height-Augmented Image}

Before converting the patch's points into the CURL representation, the patch frame $\mathtt{M}_k$ needs to be defined to produce a masked height-augmented image.
In our previous work \cite{zhang2024curl}, conformal mapping parametrization \cite{meng2016tempo} was employed for height-augmented image generation. In this paper, we propose a simplified yet efficient method that omits the quasi-conformal mapping {calculation}, thereby accelerating the optimization for odometry (Section \ref{sec:Pose Optimisation}) and BA (Section \ref{sec:localBA}). This approach also simplifies the {submap} update process (Section \ref{sec: update_sph}).


\subsubsection{{Patch Frame Definition for $\mathbf{T}_{\mathtt{L}_j\mathtt{M}_k}$}}

Three axis-aligned orthogonal reference planes can be defined at the centre of a cubic voxel, as shown in Fig. \ref{fig:projection_planes}. To select the plane that reserves the maximum information of the patch points for generating a height-augmented image, 
the normal vector $\mathbf{n}$ of the patch is calculated as
\begin{equation}
	\begin{aligned}
		 & \mathbf{n}  = \mathbf{v}_{\text{min}},\ \text{s.t.}\ \mathbf{A}\mathbf{v}_{\text{min}} = \lambda_{\text{min}} \mathbf{v}_{\text{min}}                                                                                                                                        \\
		 & \text{with}\ \mathbf{A} = \frac{1}{\psi} \sum_{i=1}^{\psi} \left( \mathbf{T}_{\mathtt{W}\mathtt{L}_j} \ {}_{\mathtt{L}_j}\mathbf{p}_{i} - \Bar{\mathbf{p}} \right) \left( \mathbf{T}_{\mathtt{W}\mathtt{L}_j} \ {}_{\mathtt{L}_j}\mathbf{p}_{i} - \Bar{\mathbf{p}} \right)^T
	\end{aligned}
	\label{eq:normal_calculation}
\end{equation}
where $\lambda_{\text{min}}$ is the minimum eigenvalue of matrix $\mathbf{A}$, $\Bar{\mathbf{p}} = \frac{1}{\psi} \sum_{i=1}^{\psi} \mathbf{T}_{\mathtt{W}\mathtt{L}_j} \ {}_{\mathtt{L}_j}\mathbf{p}_i$, and ${}_{\mathtt{L}_j}\mathbf{p}_i\in {}_{\mathtt{L}_j}\mathbf{P}^k$ denotes the $i$-th point of patch $M_k$.
The orthogonal reference plane whose normal vector is most closely aligned with $\mathbf{n}$ is then selected as the height reference plane, 
as illustrated in Fig. \ref{fig:projection_planes}.
The origin of the patch frame $\mathtt{M}_k$ in Fig. \ref{fig:patch_coordinate} is located at its centre, while its $z$ axis is perpendicular to the plane. 
Notably, we observed that when the number of patch points is small, selecting the correct height reference plane during the creation of a new {submap} patch is unstable. Therefore, instead of immediately fixing the height reference plane,   points from subsequent scans are accumulated. Once the number of points exceeds a certain threshold, we re-determine and fix the height reference plane.

\subsubsection{Weighted Height-Augmented Image with Mask}
\label{sec:Weighted Height-Augmented Image with Mask}



\begin{figure}[t]
	\centering
	\includegraphics[width=1\linewidth]{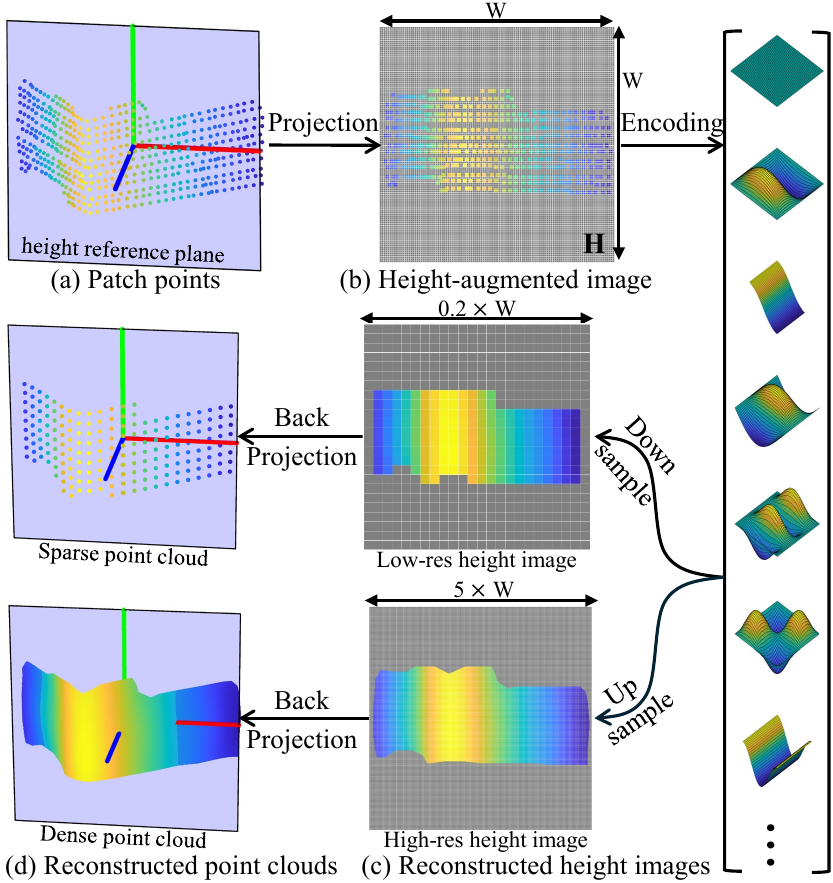}

	\caption{(a) and (b) illustrate the patch points and the corresponding height-augmented image with a resolution of $W \times W$, respectively. The grey pixels indicate the masked area. After encoding, sparser (down-sampling) and denser (up-sampling)  points can be reconstructed. Height values are color-coded using the color bar in Fig. \ref{fig:patch_coordinate}.}

	\label{fig:patch_frame}
\end{figure}

After defining 
the patch frame $\mathtt{M}_k$, a height-augmented image can be created by projecting all LiDAR points in the patch $M^k$ on to the height reference plane.
A discrete grid of a resolution of $W \times W$ is associated with the reference plane, corresponding to the width and height of the square augmented image $\mathbf{H}$, as illustrated in Fig. \ref{fig:patch_frame}(b). For simplicity, the 2D coordinates $[x, y]^T \in \mathbf{H}$ are defined in the patch frame \(\mathtt{M}_k\). {A binary mask stores} valid pixel {positions} by coding grey pixels as zero and colored {regions} as one.  

Meanwhile, a weight image $\mathbf{W}$ with the same resolution is initialized based on the range measurement $d$ in the LiDAR frame $\mathtt{L}_j$, as $e^{-2d^2/\sigma^2}$, 
where $\sigma=50$ is used in our experiment. Basically  indicates that points further away are assigned less weight. The pixels of the weight image corresponding to the invalid locations on the height image are initialized to zero. This weight image will be used to update $\mathbf{H}$ in Section \ref{sec: update_sph}.


\subsection{Spherical Harmonics Encoding}
\label{sec:spherical harmonics function encoding}
The height-augmented image $\mathbf{H}$ is encoded as spherical harmonics which is a series of spherical functions defined on the surface of a sphere {\cite{Zhang-RSS-22, shen2006large}}. 
The spherical harmonics expansion function with the maximum degree $L$ is defined as
\begin{equation}
	\small
	\begin{split}
		f(\theta,\phi) & \approx \sum_{l=0}^{L}\sum_{m = -l}^{l}c_{l,m}Y_{l,m}(\theta,\phi),\  (\theta \in[0,\pi],\phi \in[0,2\pi])
	\end{split}
	\label{eq:spherical_harmonics_expansion}
\end{equation}
where $f(\theta,\phi)$ in spherical coordinate $(\theta,\phi)$ is represented by a linear combination of the real-form spherical harmonics function $Y_{l,m}(\theta,\phi)$ with degree $l$ and order $m$, defined as
\begin{equation}
	\small
	Y_{l,m}(\theta,\phi) = \sqrt{\frac{2l+1}{4\pi} \frac{(l-|m|)!}{(l+|m|)!}}P_{l,|m|}(cos\theta)N_m(\phi)
	\label{spherical_harmonics}
\end{equation}
where {$P_{l,m}(\cdot)$} is the associated Legendre polynomials \cite{muller2006spherical}, and $N_m(\phi)$ is defined in Appendix A.
We define $\mathbf{c}=[c_{0,0},c_{1,-1}, \dots,c_{l,m},\dots,c_{L,L} ] \in \mathbb{R}^{(L+1)^2\times 1}$ as the spherical harmonics coefficients.


{To encode discrete observations using this representation, suppose we have $\varsigma$ observations sampled on the sphere. Let $\mathbf{f} = [f(\theta_1,\phi_1), \dots, f(\theta_\varsigma,\phi_\varsigma)]^\top \in \mathbb{R}^{\varsigma \times 1}$ be the vector of observed values. According to \eqref{spherical_harmonics}, we construct the spherical harmonics basis matrix $\mathbf{Y} \in \mathbb{R}^{\varsigma \times (L+1)^2}$, where each row corresponds to the evaluation of all spherical harmonics functions at one observation point. This results in the following linear system:}
{
\begin{equation}
	\label{linear_system_matrix}
	\mathbf{Y}\mathbf{c} = \mathbf{f}
\end{equation}
}
{The spherical harmonics coefficients $\mathbf{c}$ that best approximate the function values can then be obtained by solving this system via least squares.}


{To encode $\mathbf{H}$ using spherical harmonics, we first establish a bijective mapping between the coordinates $(x, y)$ in $\mathbf{H}$ and the spherical coordinates $(\theta, \phi)$. However, the north and south poles in spherical coordinates correspond to singularities, where the azimuthal angle $\phi$ becomes undefined due to all meridians converging at a single point. As a result, these poles do not preserve a one-to-one mapping. To avoid this issue and maintain bijectivity, we define the following adjusted coordinate transformation:}
\begin{equation}
	\begin{aligned}
		\theta(y) & =\frac{y-0.5s}{s}\cdot \pi \cdot \eta  +\frac{\pi}{2}\cdot(1-\eta ) \\
		\phi(x)   & =\frac{x-0.5s}{s}\cdot 2\pi \cdot \eta  +\pi\cdot(1-\eta )
	\end{aligned}
	\label{eq:mapping to spherical coordinate}
\end{equation}
where $s$ is the cubic voxel length for patch generation, and $\eta$ is a scaling factor (fixed as $0.8$ in our experiment). 
Iterative Residual Fitting method \cite{shen2006large} is used to compute the spherical harmonics coefficients for balanced accuracy and efficiency. With the spherical harmonics coefficients, continuous patch reconstructions can be achieved by reconstructing height-augmented image with a width resolution $\omega$. Fig. \ref{fig:patch_frame}(c) and \ref{fig:patch_frame}(d) show an example of continuously reconstructing the patch points in Fig. \ref{fig:patch_frame}(a) using 5-degree spherical harmonics and $\omega=0.2\times W$, as down-sampling, and $\omega=5\times W$, as up-sampling, respectively.

After finishing the spherical harmonics encoding of a patch, its coefficients $\mathbf{c}$ in the CURL representation can be used to continuously reconstruct and update the {submap}.
Next, we discuss how the CURL map can be utilized for pose estimation.

\subsection{Frame-to-Submap Pose Optimization as Odometry}
\label{sec:Pose Optimisation}


The LiDAR pose of the current scan relative to the previous scan is formulated as a frame-to-submap pose optimization problem. It minimizes the errors between the projected heights of the {raw} scan patches and the heights of their corresponding submap patches reconstructed from the CURL's spherical harmonics functions.

\begin{figure}
	\centering
	\includegraphics[width=\linewidth]{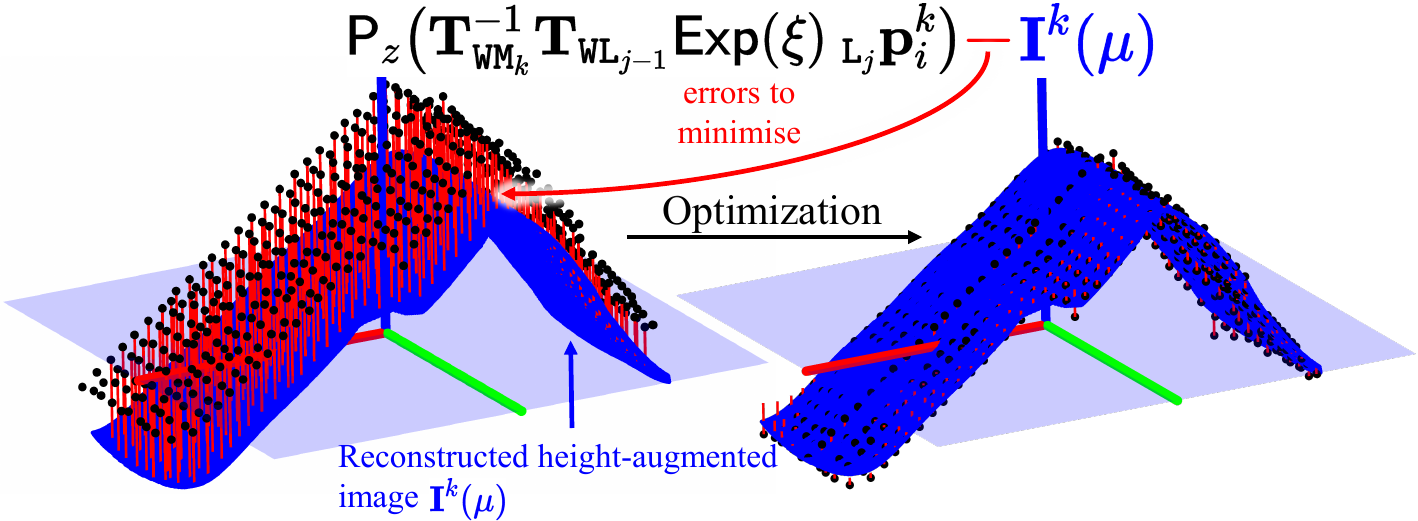}
	\caption{Pose optimization on errors between the projected heights and the reconstructed heights.}
	\label{fig:pose optimisation}
\end{figure}

Suppose there are $K$ patches in $\mathbb{M}_j$ associated with the submap patches and
${}_{\mathtt{L}_j}\mathbf{p}_i^k \in {}_{\mathtt{L}_j}\mathbf{P}^k$
denotes the $i$th point in patch $M^k$.
The objective function to be minimized is formulated as
\begin{equation}
	\begin{aligned}
		e(\bm{\xi})             & = \sum_{k=1}^{K} \sum_{i} \left | \left|
		{\mathsf{P}_z}(\mathsf{Q})
		-\mathbf{I}^k(\bm{\mu})\right | \right|^2                                                                                                                          \\
		\text{with}\ \mathsf{Q} & =\mathbf{T}_{\mathtt{W}\mathtt{M}_k}^{-1}\mathbf{T}_{\mathtt{W}\mathtt{L}_{j-1}} \mathsf{Exp}(\bm{\xi})\ {}_{\mathtt{L}_j}\mathbf{p}_i^k
	\end{aligned}
	\label{eq:residual function}
\end{equation}
where $\bm{\xi} \in \mathfrak{se}(3)$ is the Lie algebra of the 6-DoF relative pose from $\mathtt{L}_{j}$ to $\mathtt{L}_{j-1}$, $\mathsf{Exp}(\cdot)$ is the exponential map, ${\mathsf{P}_z}([x,y,z]^T)=z$, and $\mathbf{I}^k(\bm{\mu})$ is the height of the corresponding point reconstructed using the spherical harmonics:
\begin{equation}
	\begin{aligned}
		\mathbf{I}^k(\bm{\mu}) & = \sum_{l=0}^{L}\sum_{m = -l}^{l}c_{l,m}^kY_{l,m}\bigl(\theta(y),\phi(x)\bigr) \\
		\text{with}\ \bm{\mu}  & = \mathsf{P}_{xy}(\mathsf{Q})
	\end{aligned}
	\label{eq:I^k}
\end{equation}
where $c_{l,m}^k \in \mathbf{c}^k$ is the spherical harmonics coefficient of patch $M^k$, and ${\mathsf{P}_{xy}}([x,y,z]^T)=[x,y]^T$.
Note $\mathbf{T}_{\mathtt{W}\mathtt{M}_k}$ can be derived from the patch's attributes defined in Section \ref{sec:data association}. 
Hence, \eqref{eq:residual function} computes the sum of errors between the projected heights of the patches in the current LiDAR scan $j$ and the heights of the reconstructed submap patches. Intuitively, referring to Fig. \ref{fig:pose optimisation}, if the projected heights {are} considered as pixel intensities, this {is} similar to minimizing photometric errors in visual odometry and SLAM systems. \cite{kerl2013robust,forster2014svo,luo2022hybrid}.
Since $e({\bm{\xi}})$ is non-linear with respect to $\bm{\xi}$, its Jacobian matrix is required. 
The Jacobian matrix of the residual $r_{k,i}(\bm{\xi})$ inside $||\cdot||^2$ of \eqref{eq:residual function} is
\begin{equation}
	\mathbf{J} = \frac{\partial r_{k,i}(\bm{\xi})}{\partial \bm{\xi}}=
	\mathbf{J}^{\mathsf{P}_z}
	\mathbf{J}^ {\mathsf{Q}}
	- \mathbf{J}^ {\mathbf{I}^k}
	\mathbf{J}^ {\mathsf{P}_{xy}}
	\mathbf{J}^ {\mathsf{Q}}
	\label{eq:Jac}
\end{equation}
and according to  (\ref{eq:lemma1}) in Lemma 1
\begin{equation}
	\small
	\mathbf{J}^ {\mathsf{Q}} =
	\left[
		\begin{matrix}
			\mathbf{R}_{\mathtt{W}\mathtt{M}_k}^{-1}
			\mathbf{R}_{\mathtt{W}\mathtt{L}_{j-1}} \mathsf{Exp}(\bm{\theta}) &
			-\mathbf{R}_{\mathtt{W}\mathtt{M}_k}^{-1}
			\mathbf{R}_{\mathtt{W}\mathtt{L}_{j-1}} \mathsf{Exp}(\bm{\theta})[{}_{\mathtt{L}_j}\mathbf{p}_i^k]_{\times}
		\end{matrix}
		\right]
\end{equation}
{here, $\bm{\theta}$ denotes the rotational component of the $\bm{\xi}$, which is able to be written as $\bm{\xi} = [\bm{\rho}, \bm{\theta}]^\top \in \mathbb{R}^6$, where $\bm{\rho}$ represents the translational part.}
More details of these Jacobian terms are given in the Appendix A.
The Levenberg-Marquardt method is used to solve the optimization problem.

\subsection{Correction of Spherical Harmonics Coefficients}
\label{sec: update_sph}


After the pose optimization, we update the spherical harmonics coefficients of the submap patches based on the optimal pose $\bm{\xi}^*$. Using $\bm{\xi}^*$ and $\mathbf{T}_{\mathtt{WM}_k}$, we first transform the point cloud of the current scan to the corresponding patch frame. Then, following the same method described in Section \ref{sec:Weighted Height-Augmented Image with Mask}, we obtain a height-augmented image $\hat{\mathbf{H}}$ and a corresponding weight $\hat{\mathbf{W}}$. The updated height image $\mathbf{H}$ is then computed as shown in  (\ref{eq:update_H}):
\begin{equation}
	\begin{aligned}
		\mathbf{H} & = (\mathbf{H} \circ \mathbf{W} + \hat{\mathbf{H}} \circ \hat{\mathbf{W}})\oslash(\mathbf{W} + \hat{\mathbf{W}})
	\end{aligned}
	\label{eq:update_H}
\end{equation}
where $\circ$ and $\oslash$ denote element-wise multiplication and division, respectively.

	{According to \eqref{linear_system_matrix}, the vector form $\mathbf{f} = \mathbf{Y} \mathbf{c}$} allows us to recover the projected heights of a patch using its existing spherical harmonics coefficients $\mathbf{c}$. Moreover, by vectorizing $\mathbf{H}$ and removing invalid pixels, we obtain a vector $\mathbf{h}$ representing the valid height values. The residuals between the observed heights and the reconstructed heights can then be computed as:
\begin{equation}
	\mathbf{r} = \mathbf{h} - \mathbf{Y} \mathbf{c}.
	\label{eq:sph_residual}
\end{equation}
The updated spherical harmonics coefficients $\mathbf{c}^*$ can be calculated using:
\begin{equation}
	\mathbf{c}^* = \mathbf{c} + (\mathbf{Y}^T \mathbf{Y})^{-1} \mathbf{Y}^T \mathbf{r}.
	\label{eq:sph_update}
\end{equation}

Since computing the inverse of the matrix is computationally expensive and there are nearly a thousand patches to update for each scan, we accumulate data points over numerous scans for one correction
and update them simultaneously using parallel processing.
This approach significantly reduces the computational burden while maintaining real-time performance.

\section{Loop Closure and CURL-based Bundle Adjustment}\label{sec:BAandLC}

A CURL map with estimated odometry poses can be derived from the previous section. Given the map and poses, we now propose a local bundle adjustment method {that is embedded within the loop closure process, jointly optimizing the implicit spherical harmonics encoding of the map patches and the associated keyframe poses.}

Due to the inevitable error accumulation in the frame-to-submap odometry, loop closure is required to correct these errors and maintain a globally consistent map. Upon a loop closure is detected, the standard approach typically involves inserting a pose constraint, obtained from ICP alignment, into the pose graph. However, this method does not account for the consistency of other overlapping regions beyond the current frame. While a global bundle adjustment could potentially achieve a consistent map by optimizing all poses and map points simultaneously, it is computationally expensive and demands substantial resources, making it impractical for real-time applications. In this section, a local bundle adjustment is performed for the overlapping regions between the current CURL-based submap and historical CURL-based submaps. This refined information is then integrated into the global pose graph optimization, providing a balance between accuracy and computational efficiency.

\subsection{Loop Closure}
Most existing systems apply loop closure only to a pose graph, neglecting map consistency due to the substantial storage and computation requirements of point cloud maps.
To address these issues, we introduce a submap strategy in CURL-SLAM, followed by a detailed loop closure {integrated with local BA} for map consistency.

\subsubsection{Construction of the Submap Graph}
{Since all map patches are tied to specific keyframes—through an attribute that links them to the keyframe pose (to be detailed in Section \ref{sec:patch-keyframe})—we create submaps by assigning keyframe labels, where keyframes sharing the same label belong to the same submap.} When initializing a submap, the patches observed by the first keyframe of that submap are assigned the same label. For each subsequent keyframe, we monitor the number of observed patches that retain this label. If this number falls below a specified threshold (fixed as $50$ in our experiment), a new submap is created, and the new submap is recognized as a neighbor of the previous one. Only map patches belonging to the current and neighbor submaps are used as data association for pose estimation and map update. After a loop is detected and pose graph optimization is performed, the corresponding historical submaps are treated as neighbors of the current submap, enabling data association across the current and historical submaps.

\subsubsection{Loop Closure for Consistent Mapping}
Both a basic loop closure detection method and the scan context approach \cite{kim2018scan} are employed to detect loop closures. The basic loop closure method identifies historical candidates by searching within a circular region of a small radius that does not belong to its neighboring submaps. Once potential candidates are identified, we reconstruct the point cloud map from the {historical submap} patches observed by the candidate's historical keyframe.
Using the initial odometry estimation, the ICP algorithm is then applied between the current scan and the reconstructed point cloud.

\begin{figure}
	\centering
	\includegraphics[width=\linewidth]{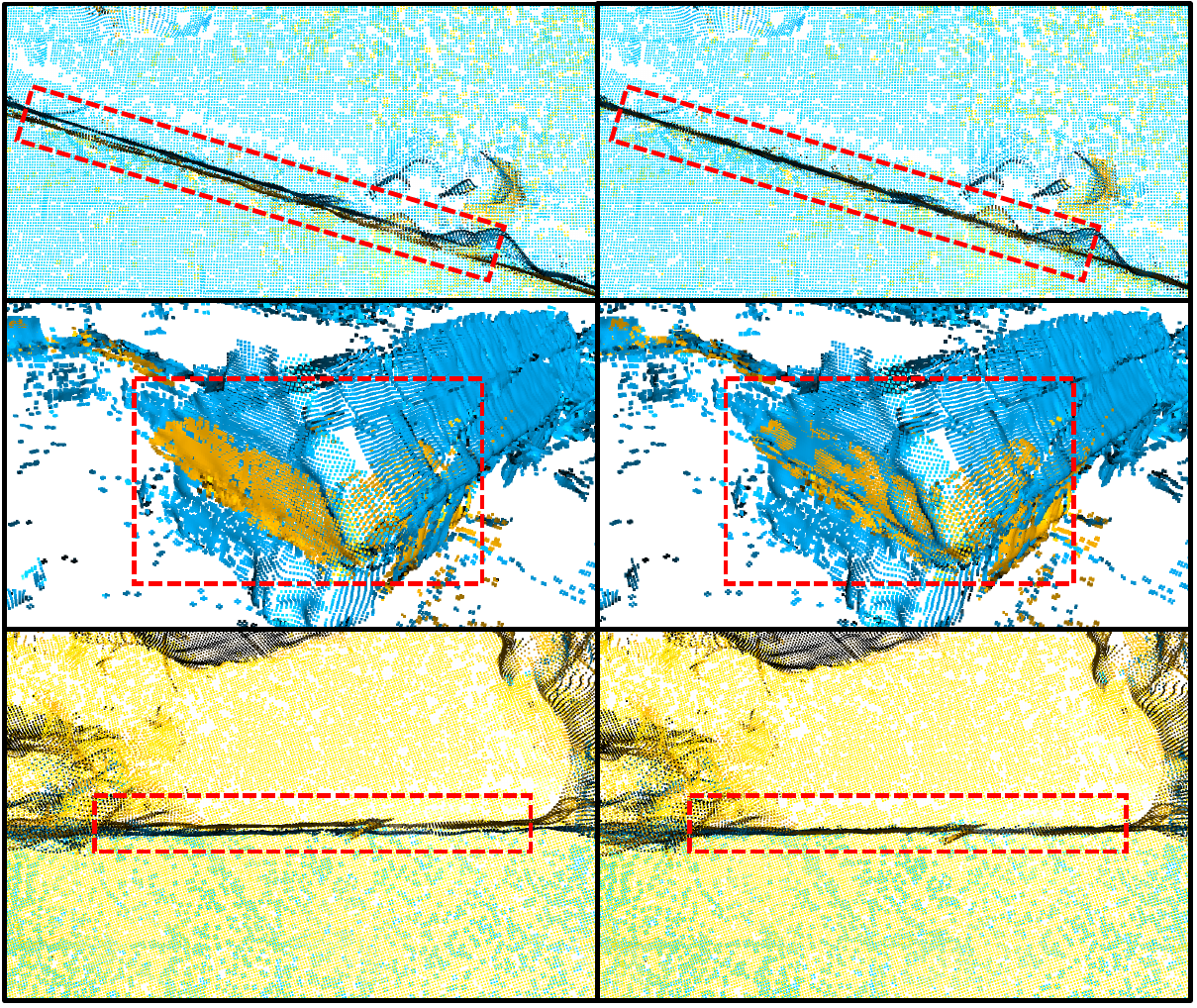}
	\caption{Reconstructed maps using pose graph optimization only (left) and the proposed local BA (right). Yellow: current submap. Blue: historical submap.}
	\label{fig:BA results}
\end{figure}

If the ICP successfully converges with an acceptable distance between the two point clouds, we consider the loop closure to be successfully detected. This constraint is then used to update the pose graph. Next, for scans from the current submap that are associated with patches from the historical submap, we perform the local BA, {which will be introduced in Section~\ref{sec:localBA}}, to optimize the keyframe trajectories of both the current and historical submaps. Once the BA is completed, these constraints are added to the pose graph for further optimization, and the patches from the current submap that overlap with those from the historical submap are merged. We can see that in Fig. \ref{fig:BA results} local BA is able to improve the map consistency.

\begin{figure}
	\centering
	\includegraphics[width=1\linewidth]{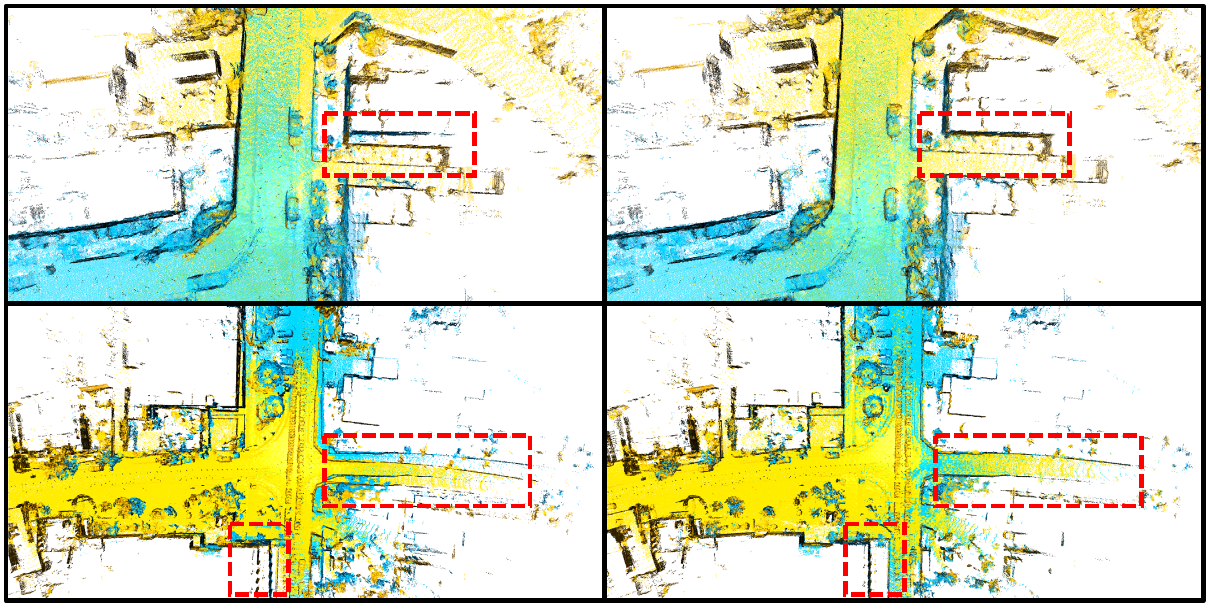}
	\caption{Reconstructed submaps before (left) and after (right) loop closure. Yellow: current submap. Blue: historical submap.}
	\label{fig:submap map merge after loop closure}
\end{figure}

Since our patches are bound to keyframe positions, once the pose graph optimization is completed, the patches' locations in $\mathtt{W}$ are updated.
Fig. \ref{fig:submap map merge after loop closure} illustrates that two submaps are consistently merged after loop closure and map update.

\subsection{Local Bundle Adjustment}\label{sec:localBA}
{We now describe the details of the local BA used in our system.}
Since our map is implicitly represented by patches' spherical harmonics functions, our CURL-based local BA is formulated to simultaneously optimize the poses and the spherical harmonics coefficients.

\subsubsection{Patch-Keyframe Association}
\label{sec:patch-keyframe}
To efficiently update the map when the pose has been refined, 
we associate patches of the map with a keyframe. As stated in Section \ref{sec:data association}, each patch has an attribute $\mathbf{T}_{\mathtt{WL}_j}$, which represents its corresponding keyframe pose. When the keyframe pose is updated, the patch's world coordinates are automatically adjusted accordingly. A new keyframe is added whenever the pose has translated a certain distance or rotated beyond a specified degree. Moreover, instead of rigidly binding the patch to its original keyframe, we dynamically associate the patch with the keyframe whose position is the nearest.

\subsubsection{BA Formulation}
Suppose all the scans of the keyframes within a local window {belong to the current submap} have been divided into {raw scan} patches. Let $\mathbb{M}_{ba}$ be a set of $K$ {raw scan} patches that have been successfully associated with a set of $N$ {historical submap} patches and are intended for use in BA. Unlike in Section \ref{sec:Pose Optimisation}, which describes a surjective relationship between the patches from the keyframe scan and the patches from the {submap}, here multiple keyframe patches may be associated with the same {submap} patch. Consequently, we have $N \leq K$.
The objective of our BA is to simultaneously optimize the keyframe poses within the local window and the spherical harmonics coefficients of the associated {submap} patches. The objective function is derived as
\begin{equation}
	\begin{aligned}
		e_{ba}(\mathbf{T}_{\mathtt{WL}_q}, \mathbf{T}_{\mathtt{WL}_j}, \mathbf{c}^n) & = \sum_{k=1}^{K} \sum_{i}\left| \left|
		{\mathsf{P}_z}(\mathsf{Q})
		-\mathbf{I}^n(\bm{\mu}) \right| \right|^2                                                                                                                                                                 \\
		\text{with}\ \mathsf{Q}                                                      & = \mathbf{T}_{\mathtt{M}_n\mathtt{L}_{q}} \mathbf{T}_{\mathtt{W}\mathtt{L}_{q}} ^{-1} \mathbf{T}_{\mathtt{W}\mathtt{L}_j}\
		{}_{\mathtt{L}_{j}}\mathbf{p}_i^k
	\end{aligned}
	\label{eq:BA}
\end{equation}
where $\mathbf{c}^n$ represents the vectorized spherical harmonics coefficients of the {submap} patch $n$, and it is encoded in $\mathbf{I}^n(\cdot)$ to reconstruct the height using \eqref{eq:I^k}.
{$\mathbf{T}_{\mathtt{WL}_q}$ denotes the pose of the keyframe $q$ with which the historical {submap} patch is associated}. {$\mathbf{T}_{\mathtt{WL}_j}$ represents the pose of the keyframe $j$ from the current submap, whose scan contains the point ${}_{\mathtt{L}_j}\mathbf{p}_i^k$.}
From (\ref{eq:BA}), we observe that only $\mathsf{Q}$ differs from  (\ref{eq:residual function}). {Alougth both of the $\mathsf{Q}$ are referring to the same concept—transforming a point from the LiDAR frame to the corresponding patch frame, the} primary difference is that we decompose $\mathbf{T}_{\mathtt{WM}_n}^{-1}$ into $\mathbf{T}_{\mathtt{M}_n \mathtt{L}_q} \mathbf{T}_{\mathtt{W}\mathtt{L}_q}^{-1}$ here. Consequently, $\mathbf{T}_{\mathtt{WL}_q}$, as well as $\mathbf{T}_{\mathtt{WL}_j}$, can be optimized during the BA.

The Jacobian matrix of the residual $r_{k,i}(\mathbf{T}_{\mathtt{WL}_q}, \mathbf{T}_{\mathtt{WL}_j}, \mathbf{c}^n)$ inside $||\cdot||^2$ of  (\ref{eq:BA}) is given by:

\begin{equation}
	\begin{aligned}
		\mathbf{J}^{r_{k,i}} = & \left[
		\mathbf{J}_{\mathbf{T}_{\mathtt{WL}_q}}^{r_{k,i}} \
		\mathbf{J}_{\mathbf{T}_{\mathtt{WL}_j}}^{r_{k,i}} \
		\mathbf{J}_{\mathbf{c}^n}^{r_{k,i}}
		\right]
		\\
		=                      &
		\mathbf{J}^{\mathsf{P}_z} \left[
		\mathbf{J}_{\mathbf{T}_{\mathtt{WL}_q}}^{\mathsf{Q}} \ \mathbf{J}_{\mathbf{T}_{\mathtt{WL}_j}}^{\mathsf{Q}} \
		\mathbf{0}
		\right] -
		\left[
		\mathbf{J}_{\mathbf{T}_{\mathtt{WL}_q}}^{\mathbf{I}^n} \
		\mathbf{J}_{\mathbf{T}_{\mathtt{WL}_j}}^{\mathbf{I}^n} \
		\mathbf{J}_{\mathbf{c}^n}^{\mathbf{I}^n}
		\right]
	\end{aligned}
	\label{eq:jacobian_residual}
\end{equation}

Here, $\mathbf{J}^{\mathsf{P}_z}$ is presented in  (\ref{eq:Z}), and according to  (\ref{eq:lemma2}) in Lemma 2, $\mathbf{J}_{\mathbf{T}_{\mathtt{WL}_q}}^{\mathsf{Q}}$ is given by:
\begin{equation}
	\small
	\mathbf{J}_{\mathbf{T}_{\mathtt{WL}_q}}^{\mathsf{Q}} =
	\left[ -\mathbf{R}_{\mathtt{M}_n\mathtt{L}_q} \quad \mathbf{R}_{\mathtt{M}_n\mathtt{L}_q} \left[ \mathbf{T}_{\mathtt{W}\mathtt{L}_q}^{-1} \mathbf{T}_{\mathtt{W}\mathtt{L}_j}\ {}_{\mathtt{L}_j}\mathbf{p}_i^k\right]_{\times}
	\right] \in \mathbb{R}^{3\times 6}
	\label{eq:J_Q_Tq}
\end{equation}
Similarly, according to  (\ref{eq:lemma1}) in Lemma 1, $\mathbf{J}_{\mathbf{T}_{\mathtt{WL}_j}}^{\mathsf{Q}}$ is given by:

\begin{equation}
	\small
	\mathbf{J}_{\mathbf{T}_{\mathtt{WL}_j}}^{\mathsf{Q}} = \left[
	\mathbf{R}_{\mathtt{M}_n\mathtt{L}_q} \mathbf{R}_{\mathtt{W}\mathtt{L}_q}^{-1} \mathbf{R}_{\mathtt{W}\mathtt{L}_j} \quad
	-\mathbf{R}_{\mathtt{M}_n\mathtt{L}_q} \mathbf{R}_{\mathtt{W}\mathtt{L}_q}^{-1} \mathbf{R}_{\mathtt{W}\mathtt{L}_j}\left[ {}_{\mathtt{L}_j}\mathbf{p}_i^k \right]_{\times}
	\right] \in \mathbb{R}^{3\times 6}
	\label{eq:J_Q_Tj}
\end{equation}

Furthermore, unlike the pose optimization \eqref{eq:Jac} in Section \ref{sec:Pose Optimisation}, which utilizes the numerical Jacobian $\mathbf{J}^{\mathbf{I}^k}$ in \eqref{eq:numerical I},
here we derive the analytical Jacobian matrix directly with respect to both the poses and the spherical harmonics coefficients for fast computation. These are denoted by $\mathbf{J}_{\mathbf{T}_{\mathtt{WL}_q}}^{\mathbf{I}^n}$, $\mathbf{J}_{\mathbf{T}_{\mathtt{WL}_j}}^{\mathbf{I}^n}$, and $\mathbf{J}_{\mathbf{c}^n}^{\mathbf{I}^n}$, allowing for the simultaneous adjustment of both keyframe {poses} and the map.
The Jacobian matrix with respect to the keyframe pose $\mathbf{T}_{\mathtt{WL}_q}$ is
\begin{equation}
	\begin{aligned}
		\mathbf{J}_{\mathbf{T}_{\mathtt{WL}_q}}^{\mathbf{I}^n} = \sum_{l=0}^{L} \sum_{m = -l}^{l} c_{l,m}^n \mathbf{J}^{Y_{l,m}} \left[
			\begin{matrix}
				\mathsf{U} \mathbf{J}^{\mathbf{Q}}_{\mathbf{T}_{\mathtt{WL}_q}} \\
				\mathsf{V} \mathbf{J}^{\mathbf{Q}}_{\mathbf{T}_{\mathtt{WL}_q}}
			\end{matrix}
			\right] \in \mathbb{R}^{1\times 6},
	\end{aligned}
	\label{eq:J_I_q}
\end{equation}
the Jacobian matrix with respect to the keyframe pose $\mathbf{T}_{\mathtt{WL}_j}$ is
\begin{equation}
	\begin{aligned}
		\mathbf{J}_{\mathbf{T}_{\mathtt{WL}_j}}^{\mathbf{I}^n} = \sum_{l=0}^{L} \sum_{m = -l}^{l} c_{l,m}^n \mathbf{J}^{Y_{l,m}} \left[
			\begin{matrix}
				\mathsf{U} \mathbf{J}^{\mathbf{Q}}_{\mathbf{T}_{\mathtt{WL}_j}} \\
				\mathsf{V} \mathbf{J}^{\mathbf{Q}}_{\mathbf{T}_{\mathtt{WL}_j}}
			\end{matrix}
			\right] \in \mathbb{R}^{1\times 6}.
	\end{aligned}
	\label{eq:J_I_j}
\end{equation}
and the Jacobian matrix with respect to the coefficients $\mathbf{c}^n$ can be easily derived based on (\ref{eq:I^k}), given by
\begin{equation}
	\mathbf{J}_{\mathbf{c}^n}^{\mathbf{I}^n} = \left[Y_{0,0}\ Y_{1,-1}\ \dots\ Y_{l,m}\ \dots\ Y_{L,L} \right]\in \mathbb{R}^{1\times (L+1)^2}
\end{equation}
Here, $\mathsf{U} = \mathbf{J}^{\phi} \mathbf{J}^{\mathsf{P}_x} \mathbf{J}^{\mathsf{P}_{xy}}$, $\mathsf{V} = \mathbf{J}^{\theta} \mathbf{J}^{\mathsf{P}_y} \mathbf{J}^{\mathsf{P}_{xy}}$, and $\mathbf{J}^{\phi}$ and $\mathbf{J}^{\theta}$ can be found in Appendix A.
$\mathbf{J}^{Y_{l,m}}$ is the most complex term, as defined in (\ref{eq:J_Y}) in Lemma 3 of Appendix B.
As illustrated in Fig. \ref{fig:jacobian of spherical harmonics function}, the tangent vector calculated by our analytical gradients (represented as red arrows) demonstrates strong alignment with the tangent vector calculated by numerical gradients (depicted as green arrows).
This alignment substantiates the correctness of our derived Jacobian matrices and the optimization process.

\begin{figure}
	\centering
	\includegraphics[width=0.7\linewidth]{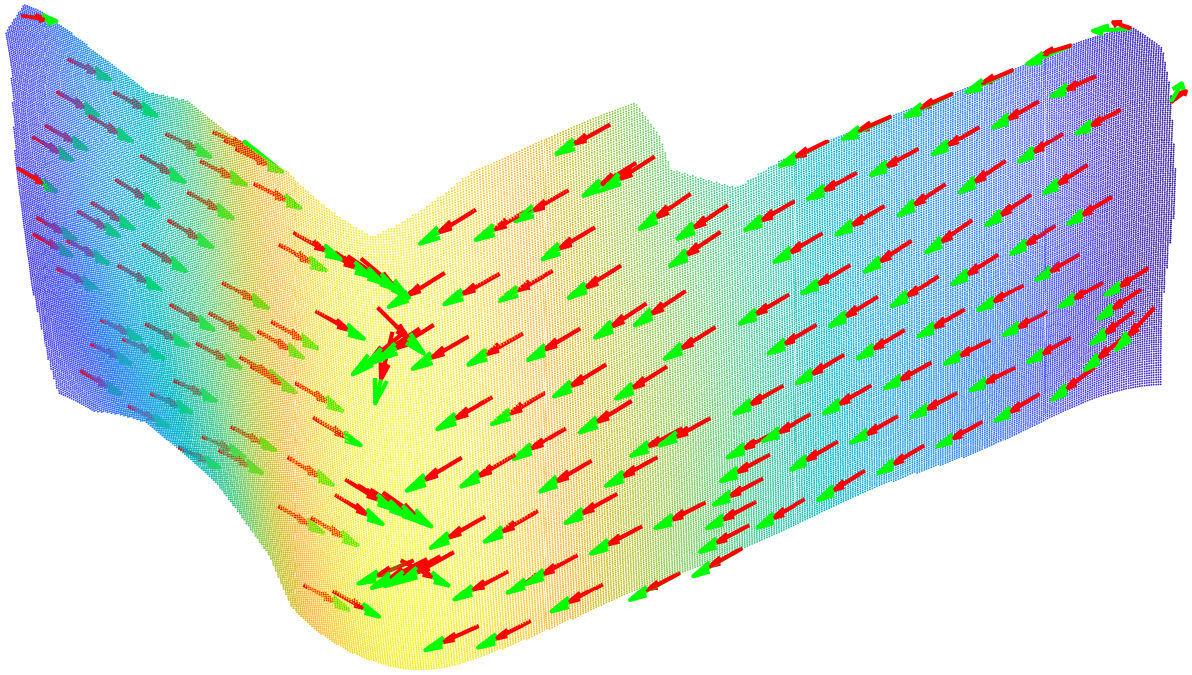}
	\caption{Comparison of numerical (green) and analytical (red) tangent vector plotted on a reconstructed point cloud.
	}
	\label{fig:jacobian of spherical harmonics function}
\end{figure}

\subsubsection{Regularization}

Although the above BA is well formulated,
in challenging cases where only a few points from a {historical submap} patch are observed in the local window scans, optimizing the spherical harmonics coefficients may degrade due to insufficient data constraints.
To address this problem, we introduce {the following} regularization term to stabilize the optimization process:
\begin{equation}
	e_{reg}(\mathbf{c}^n) = \sum_{n=1}^{N} \left \|
	\mathbf{\tilde{h}} - \mathbf{Y} \mathbf{c}^n
	\right \|^2,
	\label{eq:BA regularization term}
\end{equation}
where $\mathbf{\tilde{h}}$ represents the reconstructed height image values at valid pixels before $\mathbf{c}^n$ is optimized, and $\mathbf{Y}$ denotes the matrix of spherical harmonics basis functions.

Thus, the final objective function for the local BA is

\begin{equation}
	e = e_{ba} + \kappa e_{reg},
\end{equation}
where $\kappa = 1$ is used as the weight for the regularization term. {Again}, Fig. \ref{fig:BA results} presents the reconstructed maps using pose graph optimization only and the proposed BA. It is evident that the proposed BA significantly improves the map alignment between the current and the historical submaps by jointly optimizing the poses and the CURL coefficients, enhancing the map consistency.

\section{Implementation Details}

To achieve real-time performance on CPUs, apart from the methodology design, we carefully select appropriate data structures and introduce several strategies to accelerate our implementation tailored for the proposed CURL-SLAM.

\subsubsection{Hashing Voxel for Data Association}

Our data association method involves identifying overlapping bounding boxes from the map. However, iterating through each patch in the map can be computationally expensive. To mitigate this challenge, we adopt the approach proposed in \cite{teschner2003optimized}, which employs a spatial hashing grid in the $\mathtt{W}$ frame. This grid structure stores pointers to patches, thereby improving both memory efficiency and lookup performance.

For any 3D point $\mathbf{p}$ located within a cubic voxel with side length, $s$, a unique hash key corresponding to the voxel can be computed as
\begin{equation}
	\begin{aligned}
		\text{key}  & = [i\; j\; k]^{T} = \lceil \mathbf{p} / s \rceil,          \\
		\text{hash} & = (i \cdot P_1) \oplus (j \cdot P_2) \oplus (k \cdot P_3),
	\end{aligned}
	\label{eq:hash function}
\end{equation}
where $\lceil \rceil$ is the ceil function, $\oplus$ is bitwise XOR, and $P_1$, $P_2$, $P_3$ are large prime numbers, chosen as $73856093$, $19349669$, $83492791$ in our experiment. If the axis-aligned bounding box of a patch intersects with the voxel, a pointer to the patch's information is stored in the voxel.
By storing patch pointers in the hashing voxel, we achieve efficient access to all the CURL attributes described in Section \ref{sec:data association}. For a query axis-aligned bounding box of a patch, the hash keys for the voxel intersected with the query bounding box can be computed immediately. This allows us to efficiently retrieve the relevant map patches and use their bounding boxes to perform a more precise intersection detection, thereby enhancing both the accuracy and performance of the data association process.
Once the relevant patches are identified from the map, the insertion and removal of patches in the hashing voxel structured map are highly efficient, further reducing the computational cost of updating the map.

\subsubsection{Accelerating Odometry Performance}
Through experimentation, it is found that the reconstruction step in  (\ref{eq:I^k}) for pose optimization in odometry consumes the majority of the computational time. Additionally, the map patches observed by the current scan are highly likely to be observed in subsequent scans. Therefore, we reconstruct the height images and compute their gradients in advance. 
The precomputed reconstructed images and gradients are then efficiently updated during the patch update process, as described in Section \ref{sec: update_sph} with parallel processing.
By precomputing these values, we significantly reduce the computational overhead during pose optimization, enhancing real-time performance only using CPUs while maintaining accuracy.

\subsubsection{Managing RAM Utilization}
Although our map representation is compact, the previously mentioned precomputed reconstructed and gradient images can impose a significant memory burden during runtime. To manage memory usage effectively, we clear this information for map patches that are out of the LiDAR's line of sight and dynamically recover it when these patches re-enter the LiDAR's field of view.
This strategy ensures that memory resources are optimally allocated, maintaining system performance while reducing unnecessary memory consumption.

\subsubsection{Spherical Harmonics Lookup Table}
Repeatedly computing the values of the spherical harmonics functions $\mathbf{Y}$ for every map patch update is computationally inefficient. To address this, we precalculate a lookup table for the spherical harmonics functions, along with their corresponding Jacobian matrices, at a high resolution. This table is then stored for efficient lookup during subsequent updates.
By utilizing this precomputed lookup table, the computational cost associated with map patch update is considerably reduced, thereby improving the overall CURL-SLAM efficiency.

\section{Experiments}
We evaluate the proposed CURL-SLAM in terms of both mapping and trajectory performance. Given that the primary focus of CURL-SLAM is on mapping rather than localization, our evaluation emphasizes the mapping results.

\subsection{Setup}

\subsubsection{Datasets}
We use both the Newer College dataset \footnote{\url{https://ori-drs.github.io/newer-college-dataset/multi-cam/}} \cite{zhang2021multi} and the FusionPortable dataset \footnote{\url{https://fusionportable.github.io/dataset/fusionportable/#publications}} \cite{jiao2022fusionportable} for map evaluation due to the availability of survey-level ground truth maps obtained using a Leica BLK360. Both datasets employ 128-channel LiDAR sensors: the Newer College dataset uses an Ouster OS0-128 sensor with a 90$^\circ$ vertical field of view (FoV), while the FusionPortable dataset uses an Ouster OS1-128 sensor with a 45$^\circ$ vertical FoV.

\subsubsection{Benchmarking Methods}
In the following experiments, we present our results in comparison with SLAMesh \cite{ruan2023slamesh}, PIN-SLAM \cite{pan2024pin} and HBA \cite{liu2023large}, the state-of-the-art SLAM systems that use different mapping representations. SLAMesh, which uses a mesh-based representation, builds a mesh map online on a CPU. In contrast, PIN-SLAM incorporates a neural representation with loop closure to produce a globally consistent map, operating in real-time on certain GPUs but not on CPUs.
To evaluate with a traditional point cloud-based method, we select HBA, an offline global {hierarchical} BA method for point clouds, yielding optimal trajectory accuracy with LiDAR data. In our experiments, we use PIN-SLAM's trajectory as the input for HBA, following the finding in \cite{pan2024pin} that HBA with PIN-SLAM's trajectory performs better than the combination of HBA and MULLS \cite{pan2021mulls}.
Since the map accuracy of standard LiDAR SLAM methods, such as LOAM \cite{zhang2014loam} and LeGO-LOAM \cite{shan2018lego}, depends solely on pose accuracy, and HBA has already achieved optimal results in this regard, we exclude these methods from comparison.


For HBA, we followed the guidelines in their paper, setting the layer number to 3 for sequences with fewer than $5 \times 10^3$ poses, and to 4 for sequences with more than $5 \times 10^3$ poses. For SLAMesh, all experiments used the meshing parameters provided for the KITTI dataset, as SLAMesh does not account for LiDAR channels due to the absence of feature extraction. For PIN-SLAM, we primarily used the parameters from \texttt{run\_ncd\_128.yaml} and adjusted certain parameters for successful runs on  specific sequences, such as \texttt{cloister} and \texttt{stairs}. Additionally, we used a 20 cm voxel size for PIN-SLAM's mesh reconstruction, as described in their paper.

\subsubsection{CURL-SLAM Setting}
In our experiments, for mapping parameters the default voxel side length is set to $s=1.5\,\text{m}$ for generating both ground and non-ground patches (Section \ref{sec:patch generation}). The default degree of the spherical harmonics for ground patches is set to 2, while for non-ground patches it is set to 5, as ground patches generally exhibit less geometric variation. Additionally, the default width resolution of reconstructed height-augmented images is set to $\omega=30$ for odometry and {default} map reconstruction, with $s=1.5\,\text{m}$, resulting in a point interval on the patch of $s/\omega = 5\,\text{cm}$. The spherical harmonics coefficients of a map patch are updated using the accumulated residuals from every $5$ data associations.
The region number is set to $25$, and the data association parameters are set to $\beta_{ng} = \beta_{g} = 30$. To speed up the performance, initial scan are down-sampled using 0.2 m voxel before entering our pipeline. {All mapping parameters are kept consistent across all experiments. For pose estimation, we utilize all associated patches to enhance robustness, particularly on challenging sequences marked with an asterisk (*) in TABLE \ref{tab:ATE_trajectory}.}

All results of CURL-SLAM were conducted on a laptop equipped with an Intel i7-10875H CPU.

\subsection{Evaluation on Map Reconstruction}

\begin{figure*}[t]
	\centering
	\includegraphics[width=1\linewidth]{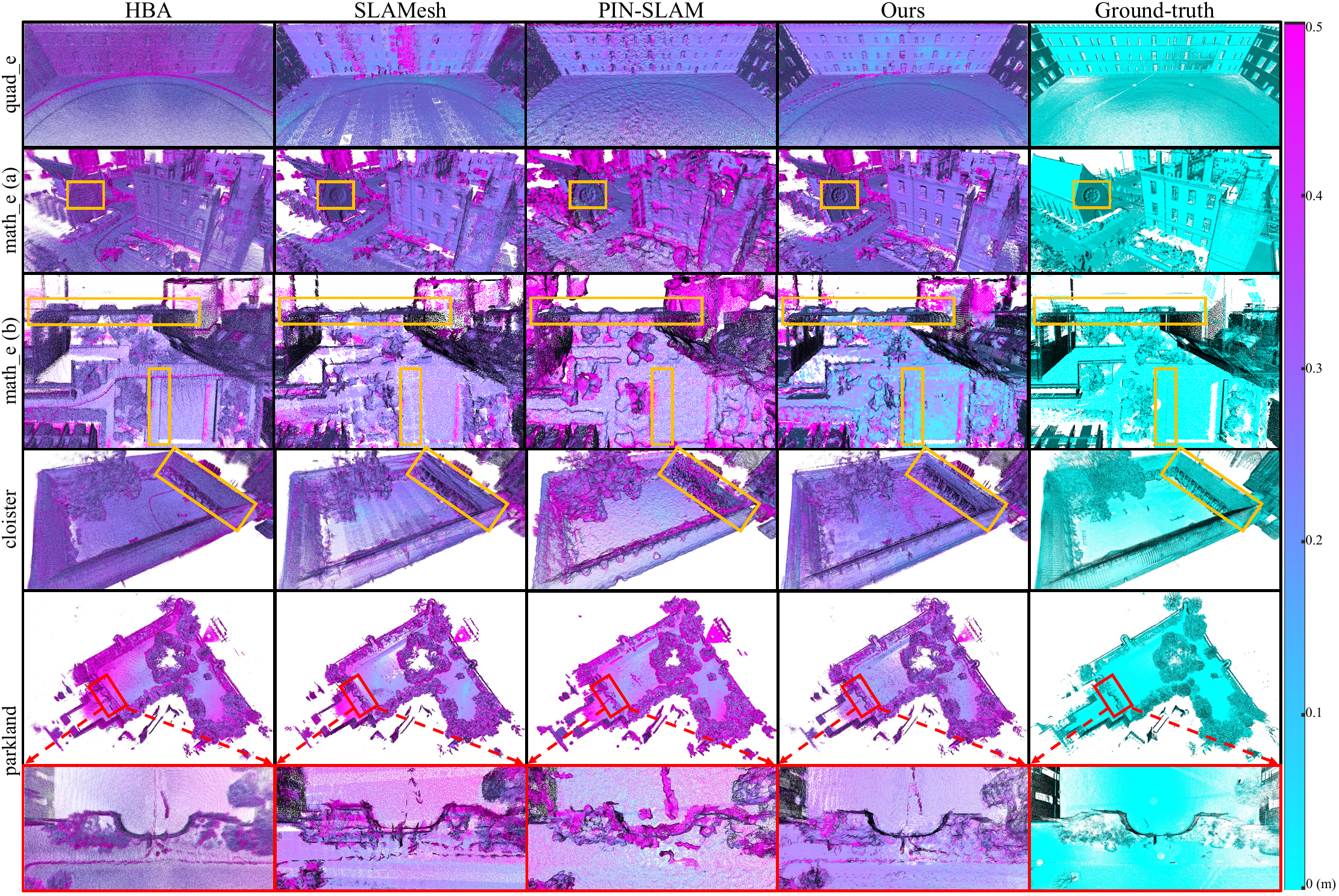}
	\caption{Reconstructed maps on the Newer College Dataset.}
	\label{fig:comparison NewerCollege}
\end{figure*}

\begin{figure*}[t]
	\centering
	\begin{subfigure}[b]{0.2\linewidth}
		\centering
		\includegraphics[width=\linewidth]{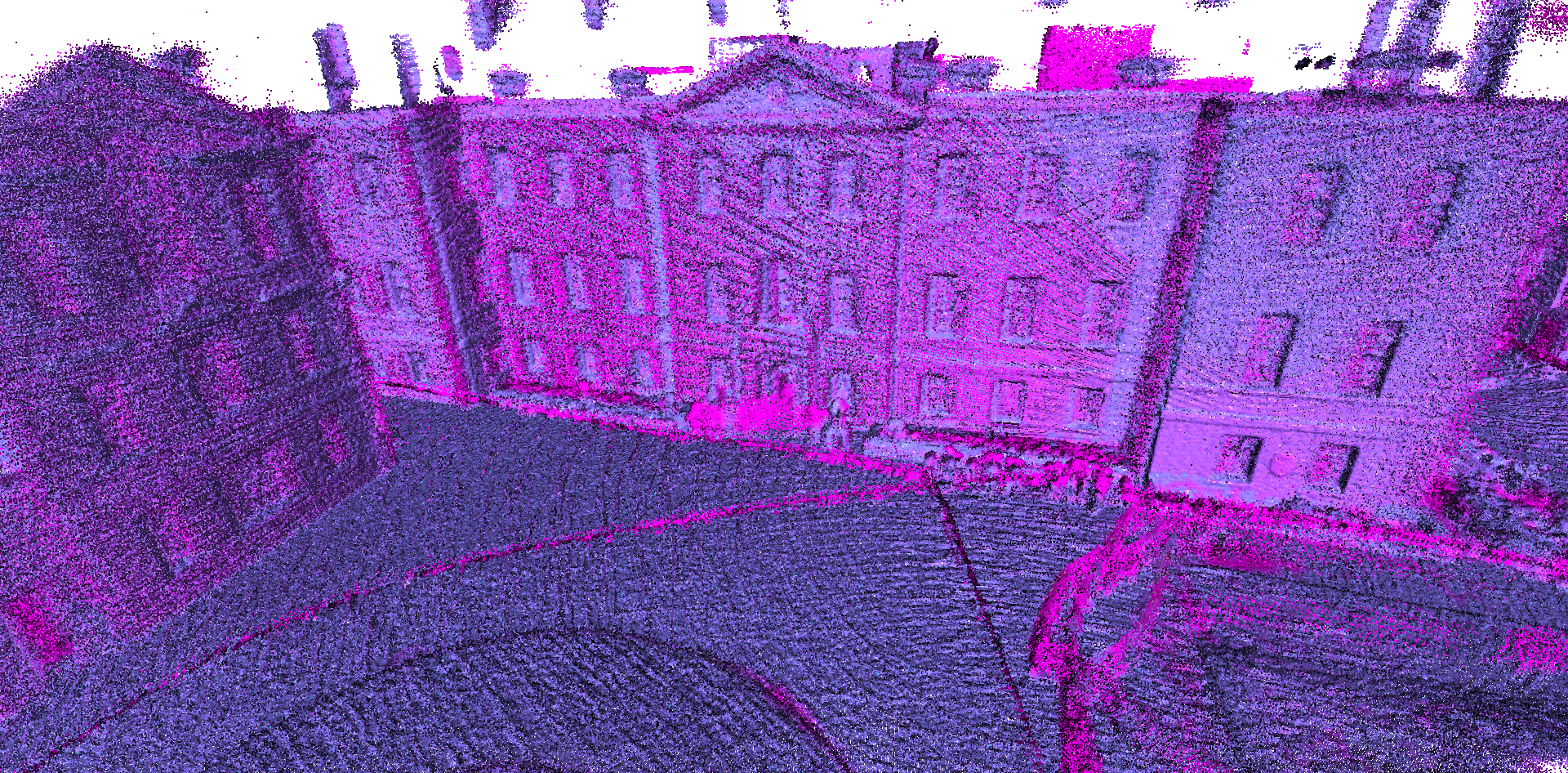}
		\caption{HBA}
	\end{subfigure}
	\begin{subfigure}[b]{0.192\linewidth}
		\centering
		\includegraphics[width=\linewidth]{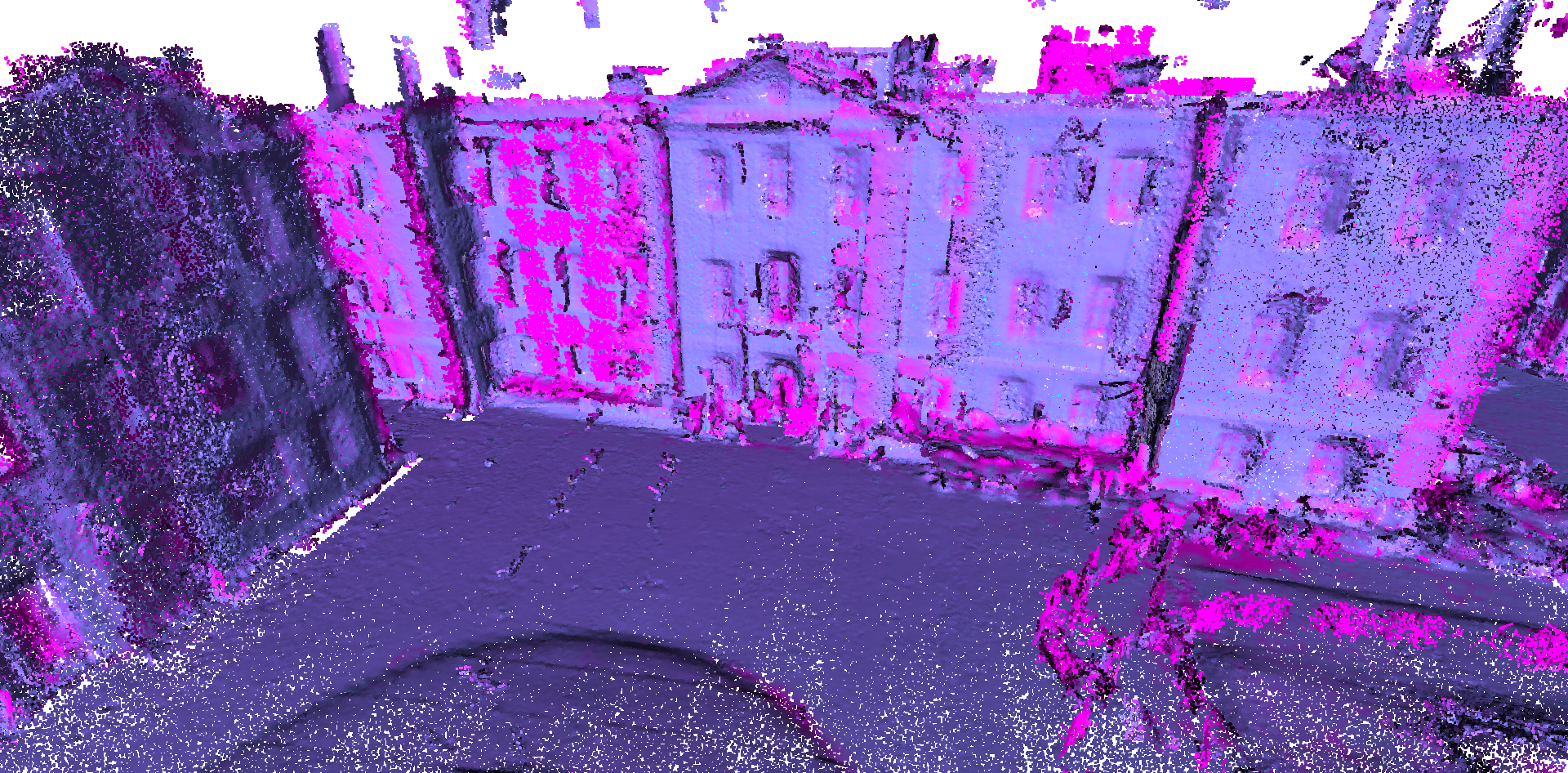}
		\caption{SLAMesh}
	\end{subfigure}
	\begin{subfigure}[b]{0.192\linewidth}
		\centering
		\includegraphics[width=\linewidth]{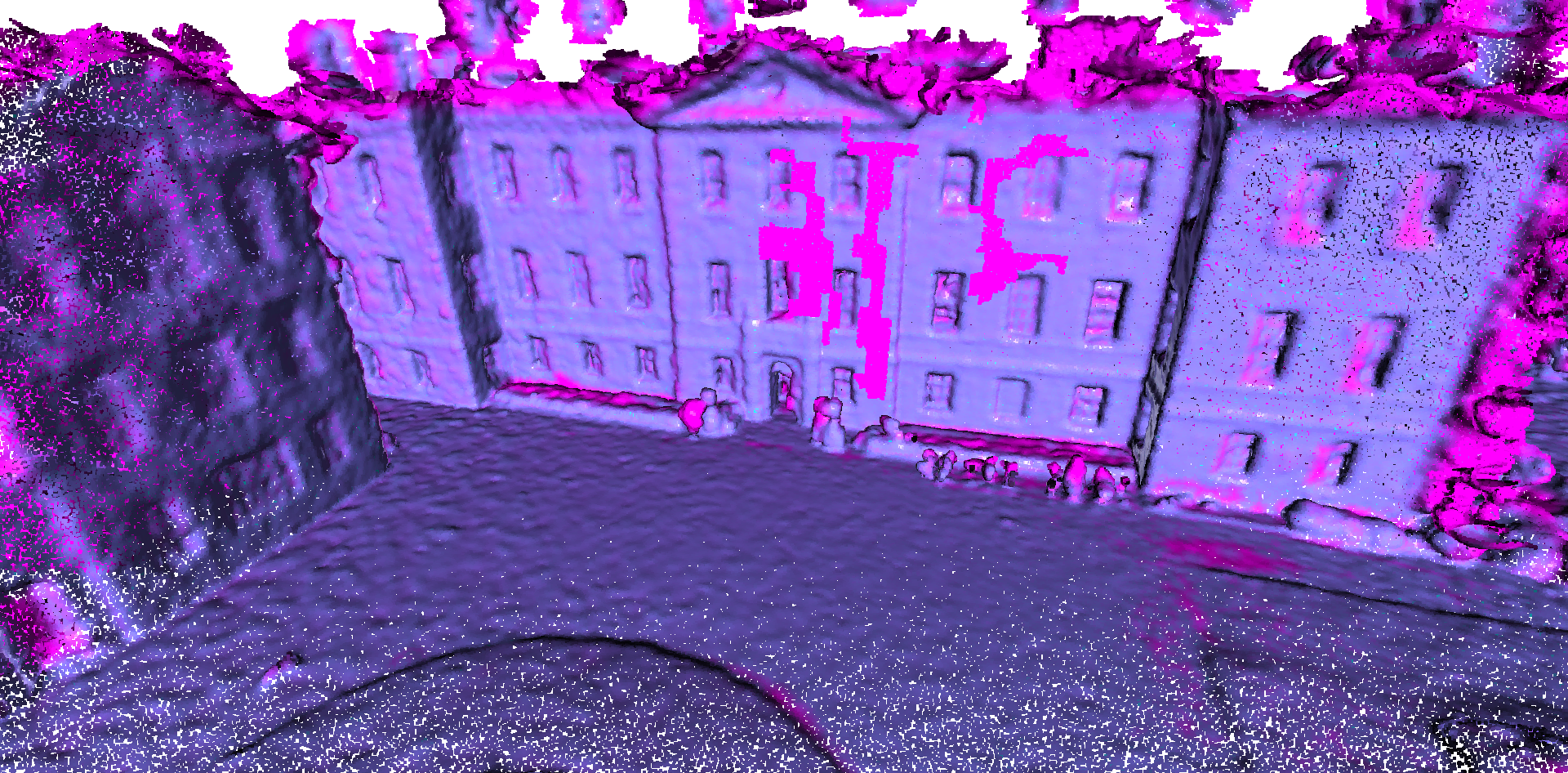}
		\caption{PIN-SLAM}
	\end{subfigure}
	\begin{subfigure}[b]{0.192\linewidth}
		\centering
		\includegraphics[width=\linewidth]{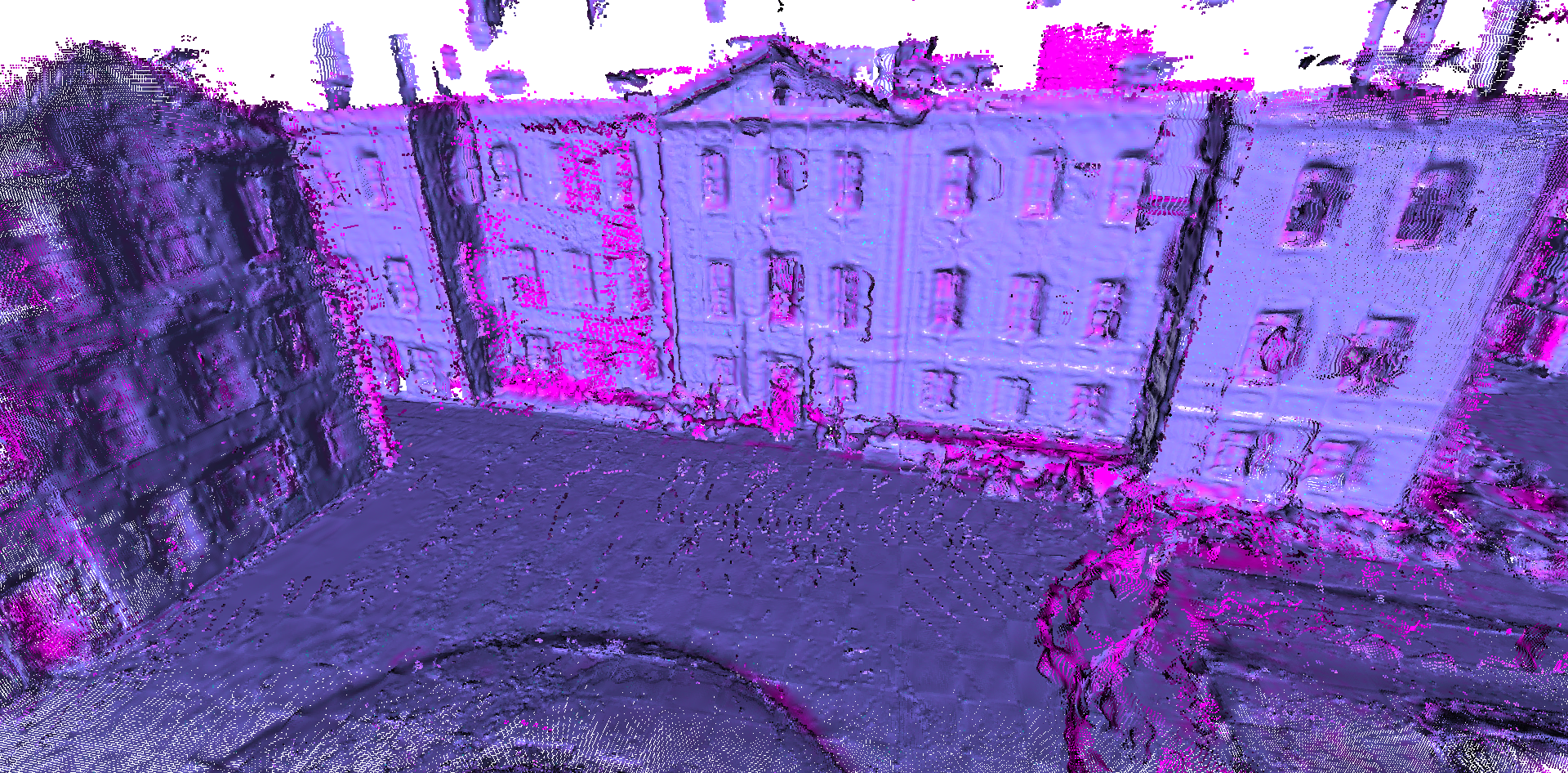}
		\caption{Ours}
	\end{subfigure}
	\begin{subfigure}[b]{0.192\linewidth}
		\centering
		\includegraphics[width=\linewidth]{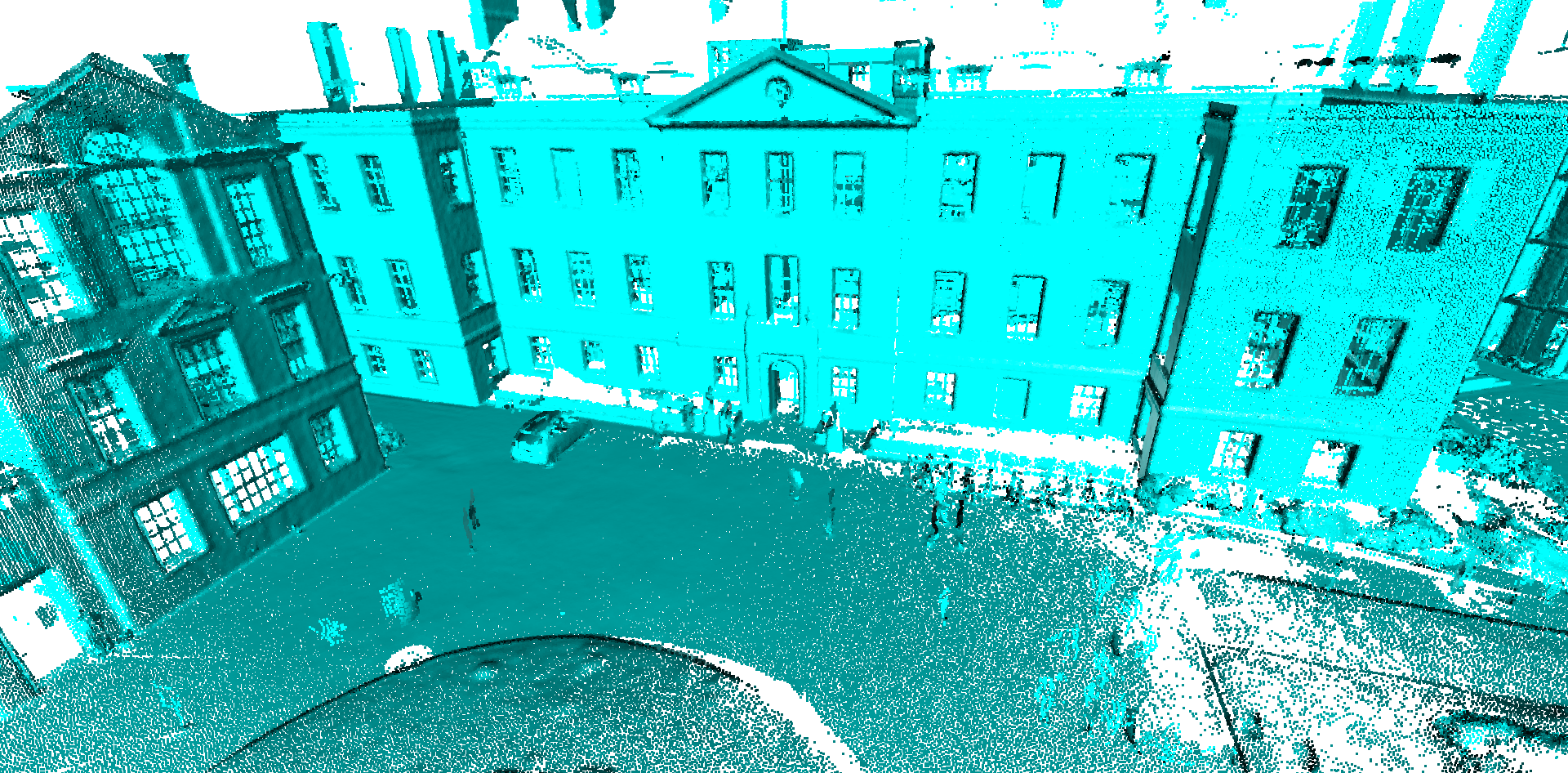}
		\caption{Ground-truth}
	\end{subfigure}
	\caption{Reconstructed maps in a region in \texttt{math\_e} scanned for a longer time and visited multiple times.}
	\label{fig:math_e_comparsion}
\end{figure*}

In this section, we present both qualitative and quantitative evaluation results on the map reconstruction. To assess the accuracy of a reconstructed map, we first align the map generated by each SLAM method with the ground truth map using ICP algorithm.

For the qualitative evaluation, we sample the same number of points from the meshes produced by SLAMesh and PIN-SLAM, as well as from our default reconstruction. The HBA point cloud map is the downsampled map using voxels with a side length of $5$ cm, matching our distance of point interval on the patch. The points are then colorized based on their distance errors to the nearest neighbor points in the ground-truth maps.

For the quantitative evaluation, both SLAMesh and PIN-SLAM generate mesh maps.
Mesh maps are built from HBA's point cloud map and from our reconstructed point cloud map, using the default resolution with the ball-pivoting algorithm \cite{bernardini1999ball}.
We then uniformly sample an equal number of points from the meshes produced by all methods. This ensures a fair comparison in terms of the map accuracy (M. Acc), map completeness (M. Comp), and L1-Chamfer distance (C-L1), as well as the F-score as a percentage (\%). Following the evaluation methodology of PIN-SLAM, M. Acc, M. Comp, and C-L1 are reported in centimeters (cm), calculated with a 20 cm error threshold.
\begin{table}
	\caption{Map evaluation results on Newer College Dataset}
	\label{tab:newer_college_results}
	\centering
	\resizebox{\linewidth}{!}{%
		\begin{tabular}{l|l|lllll}
			\hline
			\multicolumn{1}{l|}{\multirow{2}{*}{Approach}} & \multicolumn{1}{l|}{\multirow{2}{*}{Metric}} & \multicolumn{4}{c}{Sequence}                                                                                                                                                                                               \\ \cline{3-7}
			                                               &                                              & \multicolumn{1}{c|}{quad\_e}            & \multicolumn{1}{c|}{math\_e}           & \multicolumn{1}{c|}{cloister}                 & \multicolumn{1}{c|}{parkland}           & \multicolumn{1}{c}{stairs}                    \\ \hline
			\multicolumn{1}{l|}{\multirow{4}{*}{HBA}}      & M. Acc   $\downarrow$                        & \multicolumn{1}{c|}{17.83}              & \multicolumn{1}{c|}{16.15}             & \multicolumn{1}{c|}{16.32}                    & \multicolumn{1}{c|}{17.93}              & \multicolumn{1}{c}{9.88$^\dagger$}            \\ \cline{2-7}
			\multicolumn{1}{l|}{}                          & M. Comp  $\downarrow$                        & \multicolumn{1}{c|}{158.41}             & \multicolumn{1}{c|}{23.14}             & \multicolumn{1}{c|}{129.40}                   & \multicolumn{1}{c|}{113.43}             & \multicolumn{1}{c}{\textbf{186.92}$^\dagger$} \\ \cline{2-7}
			\multicolumn{1}{l|}{}                          & C-L1     $\downarrow$                        & \multicolumn{1}{c|}{88.12}              & \multicolumn{1}{c|}{19.65}             & \multicolumn{1}{c|}{72.86}                    & \multicolumn{1}{c|}{65.68}              & \multicolumn{1}{c}{\textbf{98.40}$^\dagger$}  \\ \cline{2-7}
			\multicolumn{1}{l|}{}                          & F-Score  $\uparrow$                          & \multicolumn{1}{c|}{17.83}              & \multicolumn{1}{c|}{68.06}             & \multicolumn{1}{c|}{31.11}                    & \multicolumn{1}{c|}{36.29}              & \multicolumn{1}{c}{\textbf{7.66}$^\dagger$}   \\ \hline
			\multicolumn{1}{l|}{\multirow{4}{*}{SLAMesh}}  & M. Acc   $\downarrow$                        & \multicolumn{1}{c|}{\underline{7.15}}   & \multicolumn{1}{c|}{\underline{10.75}} & \multicolumn{1}{c|}{\underline{10.80}}        & \multicolumn{1}{c|}{\underline{15.72}}  & \multicolumn{1}{c}{N/A}                       \\ \cline{2-7}
			\multicolumn{1}{l|}{}                          & M. Comp  $\downarrow$                        & \multicolumn{1}{c|}{155.73}             & \multicolumn{1}{c|}{\underline{18.52}} & \multicolumn{1}{c|}{132.22}                   & \multicolumn{1}{c|}{111.97}             & \multicolumn{1}{c}{N/A}                       \\ \cline{2-7}
			\multicolumn{1}{l|}{}                          & C-L1     $\downarrow$                        & \multicolumn{1}{c|}{\underline{81.44}}  & \multicolumn{1}{c|}{\underline{14.64}} & \multicolumn{1}{c|}{71.51}                    & \multicolumn{1}{c|}{\underline{63.85}}  & \multicolumn{1}{c}{N/A}                       \\ \cline{2-7}
			\multicolumn{1}{l|}{}                          & F-Score  $\uparrow$                          & \multicolumn{1}{c|}{\textbf{33.84}}     & \multicolumn{1}{c|}{\textbf{83.85}}    & \multicolumn{1}{c|}{39.37}                    & \multicolumn{1}{c|}{\underline{45.90}}  & \multicolumn{1}{c}{N/A}                       \\ \hline
			\multicolumn{1}{l|}{\multirow{4}{*}{PIN-SLAM}} & M. Acc   $\downarrow$                        & \multicolumn{1}{c|}{9.48}               & \multicolumn{1}{c|}{13.16}             & \multicolumn{1}{c|}{11.23/11.18*}             & \multicolumn{1}{c|}{17.39}              & \multicolumn{1}{c}{\textbf{6.04}*}            \\ \cline{2-7}
			\multicolumn{1}{l|}{}                          & M. Comp  $\downarrow$                        & \multicolumn{1}{c|}{\textbf{155.14}}    & \multicolumn{1}{c|}{19.45}             & \multicolumn{1}{c|}{\textbf{127.28}/127.64*}  & \multicolumn{1}{c|}{\underline{111.30}} & \multicolumn{1}{c}{195.01*}                   \\ \cline{2-7}
			\multicolumn{1}{l|}{}                          & C-L1     $\downarrow$                        & \multicolumn{1}{c|}{82.31}              & \multicolumn{1}{c|}{16.31}             & \multicolumn{1}{c|}{\underline{69.25}/69.41*} & \multicolumn{1}{c|}{64.34}              & \multicolumn{1}{c}{100.53*}                   \\ \cline{2-7}
			\multicolumn{1}{l|}{}                          & F-Score  $\uparrow$                          & \multicolumn{1}{c|}{32.27}              & \multicolumn{1}{c|}{75.16}             & \multicolumn{1}{c|}{\underline{42.38}/41.67*} & \multicolumn{1}{c|}{41.70}              & \multicolumn{1}{c}{4.15*}                     \\ \hline
			\multicolumn{1}{l|}{\multirow{4}{*}{Ours}}     & M. Acc   $\downarrow$                        & \multicolumn{1}{c|}{\textbf{6.46}}      & \multicolumn{1}{c|}{\textbf{9.95}}     & \multicolumn{1}{c|}{\textbf{9.95}}            & \multicolumn{1}{c|}{\textbf{13.33}}     & \multicolumn{1}{c}{\underline{6.45}}          \\ \cline{2-7}
			\multicolumn{1}{l|}{}                          & M. Comp  $\downarrow$                        & \multicolumn{1}{c|}{\underline{155.72}} & \multicolumn{1}{c|}{\textbf{18.18}}    & \multicolumn{1}{c|}{\underline{127.63}}       & \multicolumn{1}{c|}{\textbf{110.11}}    & \multicolumn{1}{c}{\underline{193.11}}        \\ \cline{2-7}
			\multicolumn{1}{l|}{}                          & C-L1     $\downarrow$                        & \multicolumn{1}{c|}{\textbf{81.09}}     & \multicolumn{1}{c|}{\textbf{14.06}}    & \multicolumn{1}{c|}{\textbf{68.79}}           & \multicolumn{1}{c|}{\textbf{61.72}}     & \multicolumn{1}{c}{\underline{99.78}}         \\ \cline{2-7}
			\multicolumn{1}{l|}{}                          & F-Score  $\uparrow$                          & \multicolumn{1}{c|}{\underline{33.29}}  & \multicolumn{1}{c|}{\underline{83.43}} & \multicolumn{1}{c|}{\textbf{42.84}}           & \multicolumn{1}{c|}{\textbf{50.16}}     & \multicolumn{1}{c}{\underline{4.77}}          \\ \hline
		\end{tabular}
	}
	\begin{flushleft}
		{\footnotesize The best results are highlighted in \textbf{bold}, and the second-best results are \underline{underlined}. N/A indicates a failure in mapping due to inconsistent regions.\\
			* Results obtained using the recommended parameters.\\
			$^\dagger$ HBA fails to produce a consistent map at this sequence although it has some good results.\\
			For the \texttt{stairs} sequence, our method used a smaller voxel side length of $s = 0.5$ m due to its narrow space.
		}
	\end{flushleft}
\end{table}

\subsubsection{Newer College}
Fig. \ref{fig:comparison NewerCollege} shows the reconstruction results from sequences of the Newer College dataset. The sequences \texttt{quad\_e}, \texttt{math\_e}, and \texttt{cloister} represent structured regions, while \texttt{parkland} demonstrates results from 2 loops in the woods from the \texttt{park} sequence, highlighting an unstructured region.

In the \texttt{quad\_e} sequence, our method clearly preserves significantly more detail compared to HBA, SLAMesh, and PIN-SLAM. HBA exhibits numerous noise points on its surface, likely due to sensor noise and scans with some levels of pose errors. SLAMesh displays blurred map details, with some white areas on the ground indicating insufficient mesh connectivity. In contrast, PIN-SLAM introduces multiple artificial bumps that do not exist in the actual environment.
Nevertheless, in this case, PIN-SLAM does demonstrate an advantage by effectively removing dynamic points.

The \texttt{math\_e (a)} further accentuates these differences. Both HBA and SLAMesh smooth out height variations, resulting in a loss of structural details. Although PIN-SLAM retains some details, it remains inferior to our method, as artificial bumps reappear on sections of the walls.

\texttt{math\_e (b)} presents a top view, showing that the wall mapped by HBA suffers from excessive thickness. 
SLAMesh displays spurious points as artifacts, while PIN-SLAM appears overly smoothed. In contrast, our method preserves greater sharpness. Additionally, ours is the only non-point-based method capable of recovering curb step edges at the bottom of the figures. Furthermore, our method accurately identifies non-observed regions, similarly to HBA, while both SLAMesh and PIN-SLAM fill gaps that should remain unfilled.

In the \texttt{cloister} sequence, only our method clearly reconstructs all the pillars, and even the texture of the roof is preserved.

The \texttt{parkland} sequence demonstrates that our method {has the potential to} perform even in unstructured environments like woods. The map produced by SLAMesh contains a ghost shadow in the red-box highlighted region, likely due to the absence of a loop closure detector and the inability to update the map for global consistency. While PIN-SLAM generates a globally consistent map, it struggles with reconstructing thin surfaces, as indicated by the substantial map errors associated. HBA also produces a globally consistent map but with noise points on its surface due to sensor noise and small errors of trajectories as mentioned before which suffers a thick and blurred surface.
In contrast, our method not only maintains global consistency but also recovers fine details with sharpness, such as the visible open gate.

The quantitative results in TABLE \ref{tab:newer_college_results} further support the observations made in Fig. \ref{fig:comparison NewerCollege}. Although the original point cloud from HBA contains many inlier points, the thick surface makes it difficult to create a well-connected mesh. Consequently, its completeness results are the worst overall. In contrast, our method does not suffer from this issue.

Additionally, for regions visited multiple times, the reconstruction of PIN-SLAM is generally good, albeit slightly smoothed like Fig. \ref{fig:math_e_comparsion} shows. However, in areas visited only once, such as some areas in the \texttt{math\_e} sequence (Figures \ref{fig:comparison NewerCollege} and \ref{fig:diff degree}), the reconstruction quality noticeably drops. This may be due to the larger number of parameters that need to be fitted for the PIN-SLAM's MLPs compared to our CURL coefficients.

\begin{figure*}[t]
	\centering
	\includegraphics[width=1\linewidth]{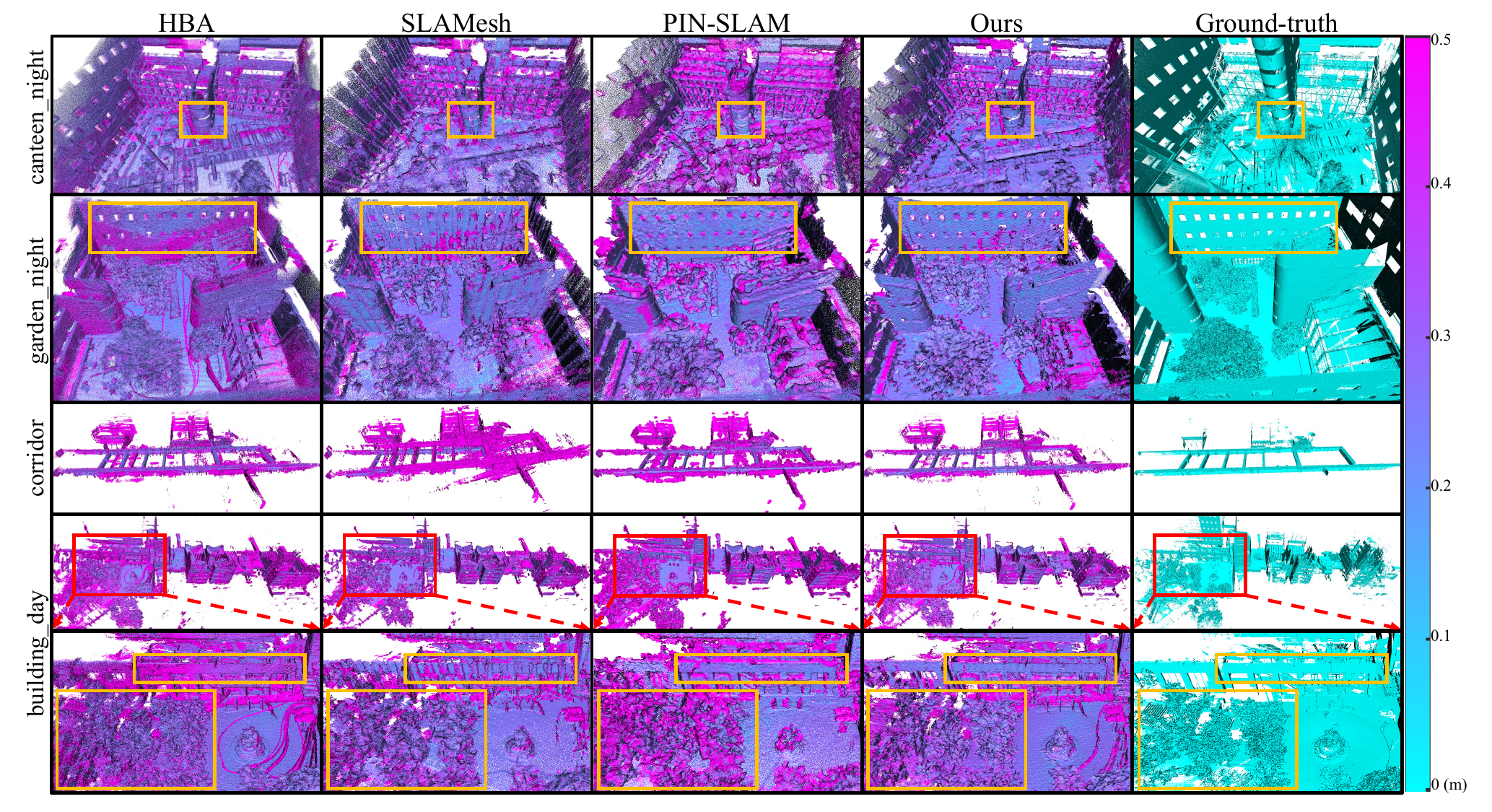}
	\caption{Reconstructed maps on the FusionPortable Dataset.}
	\label{fig:comparsion FusionPortable}
\end{figure*}

\begin{table*}[t]
	\caption{Map evaluation results on the FusionPortable Dataset}
	\label{tab:FusionPortable}
	\centering
	\renewcommand{\arraystretch}{1.3}
	\resizebox{\textwidth}{!}{%
		\begin{tabular}{l|l|cccccccc}
			\hline
			\multicolumn{1}{l|}{\multirow{2}{*}{Approach}} & \multicolumn{1}{l|}{\multirow{2}{*}{Metric}} & \multicolumn{8}{c}{Sequence}                                                                                                                                                                                                                                                                                                 \\
			\cline{3-10}
			                                               &                                              & canteen\_night                            & garden\_night                             & MCR\_fast                                & MCR\_normal                              & MCR\_slow                                & corridor\_day                          & escalator\_day                         & building\_day     \\ \hline
			\multirow{4}{*}{HBA}                           & M. Acc $\downarrow$                          & \multicolumn{1}{c|}{14.43}                & \multicolumn{1}{c|}{15.65}                & \multicolumn{1}{c|}{21.44}               & \multicolumn{1}{c|}{20.38}               & \multicolumn{1}{c|}{18.40}               & \multicolumn{1}{c|}{\underline{13.33}} & \multicolumn{1}{c|}{16.66}             & 17.72             \\ \cline{2-10}
			                                               & M. Comp $\downarrow$                         & \multicolumn{1}{c|}{78.23}                & \multicolumn{1}{c|}{46.68}                & \multicolumn{1}{c|}{31.49}               & \multicolumn{1}{c|}{27.42}               & \multicolumn{1}{c|}{25.48}               & \multicolumn{1}{c|}{\textbf{13.00}}    & \multicolumn{1}{c|}{54.15}             & 40.39             \\ \cline{2-10}
			                                               & C-L1 $\downarrow$                            & \multicolumn{1}{c|}{46.33}                & \multicolumn{1}{c|}{31.17}                & \multicolumn{1}{c|}{26.46}               & \multicolumn{1}{c|}{23.90}               & \multicolumn{1}{c|}{21.94}               & \multicolumn{1}{c|}{\textbf{13.16}}    & \multicolumn{1}{c|}{35.41}             & 29.05             \\ \cline{2-10}
			                                               & F-Score $\uparrow$                           & \multicolumn{1}{c|}{55.78}                & \multicolumn{1}{c|}{48.74}                & \multicolumn{1}{c|}{31.79}               & \multicolumn{1}{c|}{36.88}               & \multicolumn{1}{c|}{45.63}               & \multicolumn{1}{c|}{\textbf{80.27}}    & \multicolumn{1}{c|}{51.83}             & 53.21             \\ \hline
			\multirow{4}{*}{SLAMesh}                       & M. Acc $\downarrow$                          & \multicolumn{1}{c|}{\underline{10.65}}    & \multicolumn{1}{c|}{\underline{8.32}}     & \multicolumn{1}{c|}{10.53}               & \multicolumn{1}{c|}{9.26}                & \multicolumn{1}{c|}{9.36}                & \multicolumn{1}{c|}{N/A}               & \multicolumn{1}{c|}{\textbf{11.24}}    & \textbf{12.67}    \\ \cline{2-10}
			                                               & M. Comp $\downarrow$                         & \multicolumn{1}{c|}{\underline{73.18}}    & \multicolumn{1}{c|}{\textbf{34.67}}       & \multicolumn{1}{c|}{\textbf{5.68}}       & \multicolumn{1}{c|}{\textbf{5.84}}       & \multicolumn{1}{c|}{\textbf{6.04}}       & \multicolumn{1}{c|}{N/A}               & \multicolumn{1}{c|}{\textbf{43.51}}    & \underline{34.39} \\ \cline{2-10}
			                                               & C-L1 $\downarrow$                            & \multicolumn{1}{c|}{\underline{41.91}}    & \multicolumn{1}{c|}{\textbf{21.50}}       & \multicolumn{1}{c|}{\textbf{8.11}}       & \multicolumn{1}{c|}{\textbf{7.55}}       & \multicolumn{1}{c|}{\textbf{7.70}}       & \multicolumn{1}{c|}{N/A}               & \multicolumn{1}{c|}{\textbf{27.38}}    & \underline{23.53} \\ \cline{2-10}
			                                               & F-Score $\uparrow$                           & \multicolumn{1}{c|}{\textbf{69.38}}       & \multicolumn{1}{c|}{\textbf{83.13}}       & \multicolumn{1}{c|}{87.69}               & \multicolumn{1}{c|}{\textbf{90.14}}      & \multicolumn{1}{c|}{\underline{89.62}}   & \multicolumn{1}{c|}{N/A}               & \multicolumn{1}{c|}{\textbf{78.01}}    & \underline{72.57} \\ \hline
			\multirow{4}{*}{PIN-SLAM}                      & M. Acc $\downarrow$                          & \multicolumn{1}{c|}{12.65}                & \multicolumn{1}{c|}{9.69}                 & \multicolumn{1}{c|}{\textbf{9.42}}       & \multicolumn{1}{c|}{\underline{8.71}}    & \multicolumn{1}{c|}{\underline{8.66}}    & \multicolumn{1}{c|}{13.99}             & \multicolumn{1}{c|}{13.92}             & 15.43             \\ \cline{2-10}
			                                               & M. Comp $\downarrow$                         & \multicolumn{1}{c|}{\textbf{72.80}}       & \multicolumn{1}{c|}{36.22}                & \multicolumn{1}{c|}{{\underline {7.05}}} & \multicolumn{1}{c|}{{\underline{6.79}}}  & \multicolumn{1}{c|}{{\underline {7.04}}} & \multicolumn{1}{c|}{\underline{13.68}} & \multicolumn{1}{c|}{46.18}             & 35.91             \\ \cline{2-10}
			                                               & C-L1 $\downarrow$                            & \multicolumn{1}{c|}{42.73}                & \multicolumn{1}{c|}{22.96}                & \multicolumn{1}{c|}{\underline{8.23}}    & \multicolumn{1}{c|}{\underline{7.75}}    & \multicolumn{1}{c|}{7.85}                & \multicolumn{1}{c|}{\underline{13.83}} & \multicolumn{1}{c|}{30.05}             & 25.67             \\ \cline{2-10}
			                                               & F-Score $\uparrow$                           & \multicolumn{1}{c|}{63.51}                & \multicolumn{1}{c|}{76.94}                & \multicolumn{1}{c|}{{\textbf {88.34}}}   & \multicolumn{1}{c|}{89.22}               & \multicolumn{1}{c|}{89.25}               & \multicolumn{1}{c|}{76.70}             & \multicolumn{1}{c|}{68.25}             & {65.38}           \\ \hline
			\multirow{4}{*}{Ours}                          & M. Acc $\downarrow$                          & \multicolumn{1}{c|}{{\textbf {9.92}}}     & \multicolumn{1}{c|}{\textbf{7.37}}        & \multicolumn{1}{c|}{{\underline {9.48}}} & \multicolumn{1}{c|}{{\textbf{8.69}}}     & \multicolumn{1}{c|}{{\textbf{8.13}}}     & \multicolumn{1}{c|}{\textbf{12.24}}    & \multicolumn{1}{c|}{\underline{11.92}} & \underline{13.19} \\ \cline{2-10}
			                                               & M. Comp $\downarrow$                         & \multicolumn{1}{c|}{73.67}                & \multicolumn{1}{c|}{{\underline {35.69}}} & \multicolumn{1}{c|}{8.03}                & \multicolumn{1}{c|}{7.62}                & \multicolumn{1}{c|}{7.43}                & \multicolumn{1}{c|}{ 16.87}            & \multicolumn{1}{c|}{\underline{45.30}} & \textbf{31.95}    \\ \cline{2-10}
			                                               & C-L1 $\downarrow$                            & \multicolumn{1}{c|}{{\textbf{41.80}}}     & \multicolumn{1}{c|}{{\underline {21.53}}} & \multicolumn{1}{c|}{8.75}                & \multicolumn{1}{c|}{8.16}                & \multicolumn{1}{c|}{{\underline {7.78}}} & \multicolumn{1}{c|}{{ 14.56}}          & \multicolumn{1}{c|}{\underline{28.61}} & \textbf{22.57}    \\ \cline{2-10}
			                                               & F-Score $\uparrow$                           & \multicolumn{1}{c|}{{\underline {68.78}}} & \multicolumn{1}{c|}{{\underline {81.31}}} & \multicolumn{1}{c|}{\underline{88.27}}   & \multicolumn{1}{c|}{{\underline{89.62}}} & \multicolumn{1}{c|}{{\textbf{90.49}}}    & \multicolumn{1}{c|}{\underline{77.53}} & \multicolumn{1}{c|}{\underline{75.94}} & \textbf{72.90}    \\ \hline
		\end{tabular}
	}
	\begin{flushleft}
		{\footnotesize The best results are highlighted in \textbf{bold}, while the second-best results are \underline{underlined}. N/A indicates a failure in mapping due to inconsistent regions.}
	\end{flushleft}
\end{table*}

\subsubsection{FusionPortable}
The 3D maps reconstructed by the methods on the FusionPortable dataset are shown in Fig. \ref{fig:comparsion FusionPortable}.
From the reconstructions of the \texttt{canteen\_night} and \texttt{garden\_night} sequences, it is clear that our method more accurately identifies most of the unoccupied regions, as well as the point cloud map from the HBA method.
In the \texttt{garden\_night} sequence, it is evident that SLAMesh fails to reconstruct a smooth building wall. Additionally, SLAMesh cannot produce a consistent map in the \texttt{corridor} sequence due to the lack of loop closure, whereas both PIN-SLAM and our method successfully reconstruct consistent maps.

For the \texttt{building\_day} sequence, although SLAMesh achieves the best accuracy according to TABLE \ref{tab:FusionPortable}, its reconstruction of the wall is less smooth compared to PIN-SLAM and our method. HBA also produces noisy points on the wall. PIN-SLAM, on the other hand, exhibits larger errors in the woodland area.

According to the quantitative results in TABLE \ref{tab:FusionPortable}, our method achieves the highest map reconstruction accuracy across most sequences. Even when it ranks second, its map accuracy is only marginally lower than the best results.

	{Overall, CURL-SLAM consistently outperforms existing methods across both the Newer College and FusionPortable datasets, achieving the best or second-best performance in nearly all map evaluation metrics. It demonstrates strong capabilities in reconstructing accurate, complete and sharp maps—especially in various environments. 
	}

\subsection{Evaluation on Continuous Map Reconstruction}
\begin{figure*}[t]
	\centering
	\includegraphics[width=1\linewidth]{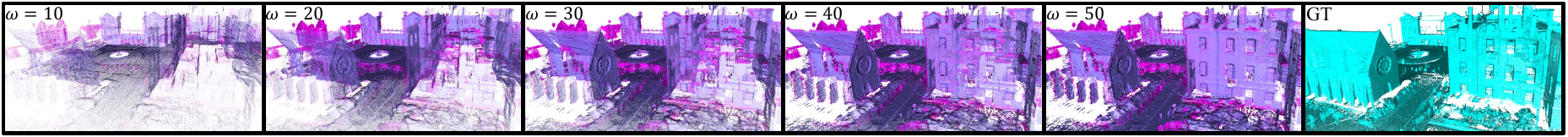}
	\caption{Continuous reconstruction of CURL-SLAM with $\omega$ values corresponding to TABLE \ref{tab:continuous reconstruction}(a). }
	\label{fig:continuous reconstruction}
\end{figure*}

\begin{table}[!htb]
	\caption{Continuous reconstruction on \texttt{math\_e}} \label{tab:continuous reconstruction}
	\begin{subtable}{.572\linewidth}
		\caption{CURL-SLAM}
		\label{tab:continuous reconstruction CURL}
		\centering
		\resizebox{\linewidth}{!}{
			\begin{tabular}{l|ccccc}
				\hline
				\multicolumn{1}{l|}{\multirow{2}{*}{Metric}} & \multicolumn{5}{c}{$\omega$}                                                                                                                                                                                                                                                                                                                                                                                                                                                                                \\ \cline{2-6}
				                                             & \multicolumn{1}{c|}{10}                                                                              & \multicolumn{1}{c|}{20}                                                                              & \multicolumn{1}{c|}{30}                                                                              & \multicolumn{1}{c|}{40}                                                                              & 50                                                                              \\ \hline
				M. Acc   $\downarrow$                        & \multicolumn{1}{c|}{\cellcolor[rgb]{0.573283950617284,	0.686543209876543,	0.304839506172839}{9.53}}  & \multicolumn{1}{c|}{\cellcolor[rgb]{0.57,	0.69,	0.3}{9.46}}                                          & \multicolumn{1}{c|}{\cellcolor[rgb]{0.571876543209877,	0.688024691358025,	0.302765432098765}{9.50}}  & \multicolumn{1}{c|}{\cellcolor[rgb]{0.571876543209877,	0.688024691358025,	0.302765432098765}{9.50}}  & \cellcolor[rgb]{0.573283950617284,	0.686543209876543,	0.304839506172839}{9.53}  \\ \hline
				M. Comp  $\downarrow$                        & \multicolumn{1}{c|}{\cellcolor[rgb]{0.760000000000000,	0.490000000000000,	0.580000000000000}{23.13}} & \multicolumn{1}{c|}{\cellcolor[rgb]{0.647520000000000,	0.608400000000000,	0.414240000000000}{20.17}} & \multicolumn{1}{c|}{\cellcolor[rgb]{0.611040000000000,	0.646800000000000,	0.360480000000000}{19.21}} & \multicolumn{1}{c|}{\cellcolor[rgb]{0.590900000000000,	0.668000000000000,	0.330800000000000}{18.68}} & \cellcolor[rgb]{0.577980000000000,	0.681600000000000,	0.311760000000000}{18.34} \\ \hline
				C-L1     $\downarrow$                        & \multicolumn{1}{c|}{\cellcolor[rgb]{0.724982935153584,	0.526860068259386,	0.528395904436860}{16.33}} & \multicolumn{1}{c|}{\cellcolor[rgb]{0.627064846416382,	0.629931740614334,	0.384095563139932}{14.82}} & \multicolumn{1}{c|}{\cellcolor[rgb]{0.596587030716724,	0.662013651877133,	0.339180887372014}{14.35}} & \multicolumn{1}{c|}{\cellcolor[rgb]{0.579726962457338,	0.679761092150171,	0.314334470989761}{14.09}} & \cellcolor[rgb]{0.570000000000000,	0.690000000000000,	0.300000000000000}{13.94} \\ \hline
				F-Score  $\uparrow$                          & \multicolumn{1}{c|}{\cellcolor[rgb]{0.667387978142076,	0.587486338797814,	0.443519125683060}{79.51}} & \multicolumn{1}{c|}{\cellcolor[rgb]{0.593256830601093,	0.665519125683060,	0.334273224043716}{83.08}} & \multicolumn{1}{c|}{\cellcolor[rgb]{0.577683060109290,	0.681912568306011,	0.311322404371585}{83.83}} & \multicolumn{1}{c|}{\cellcolor[rgb]{0.572284153005464,	0.687595628415301,	0.303366120218579}{84.09}} & \cellcolor[rgb]{0.570000000000000,	0.690000000000000,	0.300000000000000}{84.20} \\ \hline
			\end{tabular}
		}
	\end{subtable}
	\begin{subtable}{.42\linewidth}
		\caption{PIN-SLAM}
		\label{tab:continuous reconstruction PIN-SLAM}
		\centering
		\resizebox{\linewidth}{!}{
			\begin{tabular}{l|ccc}
				\hline
				\multicolumn{1}{l|}{\multirow{2}{*}{Metric}} & \multicolumn{3}{c}{Voxel Size}                                                                                                                                                                                                                                                                                    \\ \cline{2-4}
				                                             & \multicolumn{1}{c|}{24 cm}                                                                           & \multicolumn{1}{c|}{20 cm}                                                                           & \multicolumn{1}{c}{5 cm}                                                                            \\ \hline
				M. Acc   $\downarrow$                        & \multicolumn{1}{c|}{\cellcolor[rgb]{0.731382716049383,	0.520123456790123,	0.537827160493827}{12.90}} & \multicolumn{1}{c|}{\cellcolor[rgb]{0.744049382716049,	0.506790123456790,	0.556493827160494}{13.17}} & \multicolumn{1}{c}{\cellcolor[rgb]{0.76,	0.49,	0.58}{13.51}}                                        \\ \hline
				M. Comp  $\downarrow$                        & \multicolumn{1}{c|}{\cellcolor[rgb]{0.672980000000000,	0.581600000000000,	0.451760000000000}{20.84}} & \multicolumn{1}{c|}{\cellcolor[rgb]{0.620160000000000,	0.637200000000000,	0.373920000000000}{19.45}} & \multicolumn{1}{c}{\cellcolor[rgb]{0.570000000000000,	0.690000000000000,	0.300000000000000}{18.13}} \\ \hline
				C-L1     $\downarrow$                        & \multicolumn{1}{c|}{\cellcolor[rgb]{0.760000000000000,	0.490000000000000,	0.580000000000000}{16.87}} & \multicolumn{1}{c|}{\cellcolor[rgb]{0.723686006825938,	0.528225255972696,	0.526484641638225}{16.31}} & \multicolumn{1}{c}{\cellcolor[rgb]{0.691911262798635,	0.561672354948805,	0.479658703071672}{15.82}} \\ \hline
				F-Score  $\uparrow$                          & \multicolumn{1}{c|}{\cellcolor[rgb]{0.760000000000000,	0.490000000000000,	0.580000000000000}{75.05}} & \multicolumn{1}{c|}{\cellcolor[rgb]{0.757923497267759,	0.492185792349727,	0.576939890710382}{75.15}} & \multicolumn{1}{c}{\cellcolor[rgb]{0.745672131147541,	0.505081967213115,	0.558885245901639}{75.74}} \\ \hline
			\end{tabular}
		}
	\end{subtable}
	\begin{flushleft}
		{\footnotesize * The greener, the better result.}
	\end{flushleft}
\end{table}

The proposed CURL-SLAM method is capable of continuous reconstruction by adjusting the parameter $\omega$, as detailed in Section \ref{sec:spherical harmonics function encoding}. To evaluate our results on continuous reconstruction, we use the ball-pivoting algorithm to generate meshes from maps reconstructed with increasing resolutions.  TABLE \ref{tab:continuous reconstruction}(a) shows that, even when reconstructed at different resolutions, the accuracy of our map remains relatively consistent. This indicates that our continuous reconstruction avoids over-fitting or over-sampling, ensuring map reconstruction reflects true 3D scenes and structures.
Additionally, as $\omega$ increases, our completeness, L1-Chamfer distance, and F-score improve, which is expected. This indicates that our map covers more regions of the ground-truth map. The reconstructed maps corresponding to different $\omega$ settings in TABLE \ref{tab:continuous reconstruction}(a) are visualized in Fig. \ref{fig:continuous reconstruction}.

\begin{figure}[t]
	\centering
	\begin{subfigure}[b]{0.49\linewidth}
		\centering
		\includegraphics[width=\linewidth]{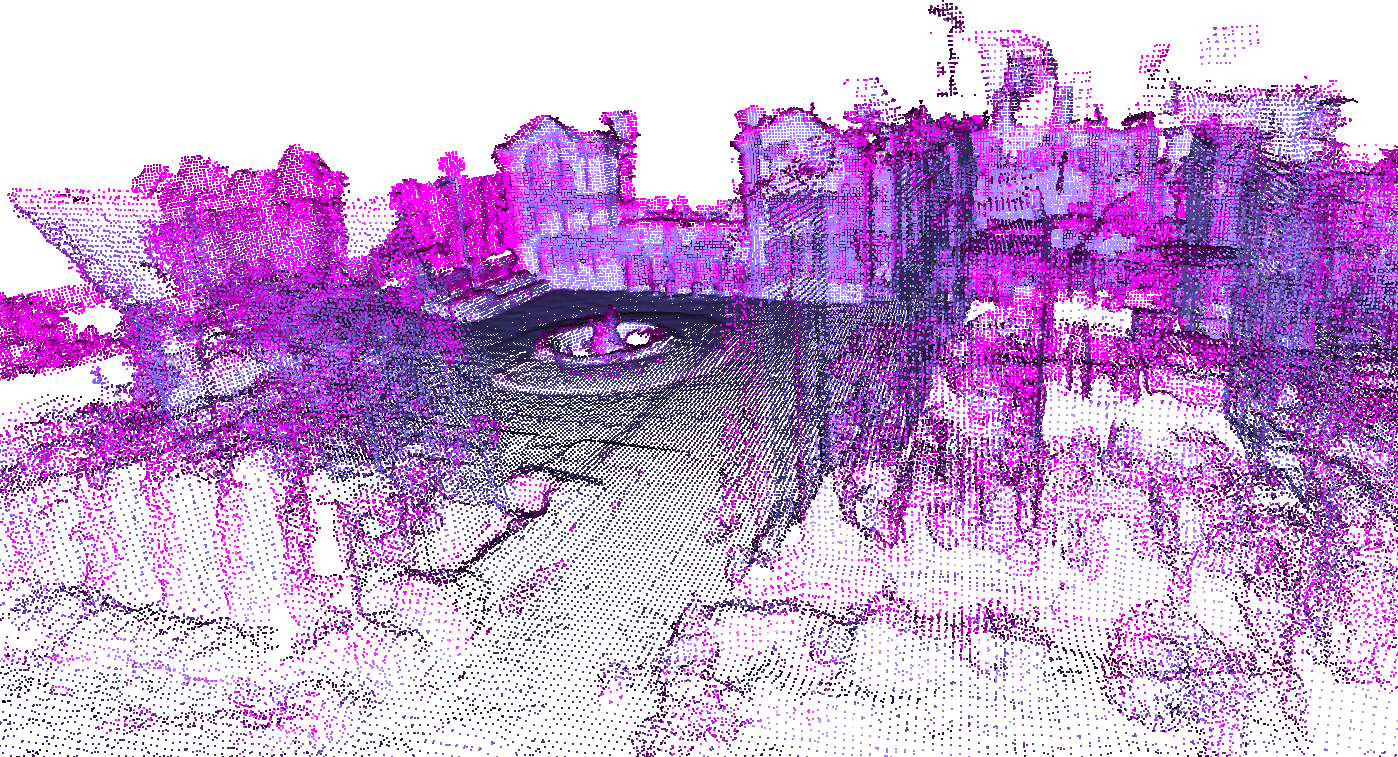}
		\caption{24 cm}
	\end{subfigure}
	\begin{subfigure}[b]{0.49\linewidth}
		\centering
		\includegraphics[width=\linewidth]{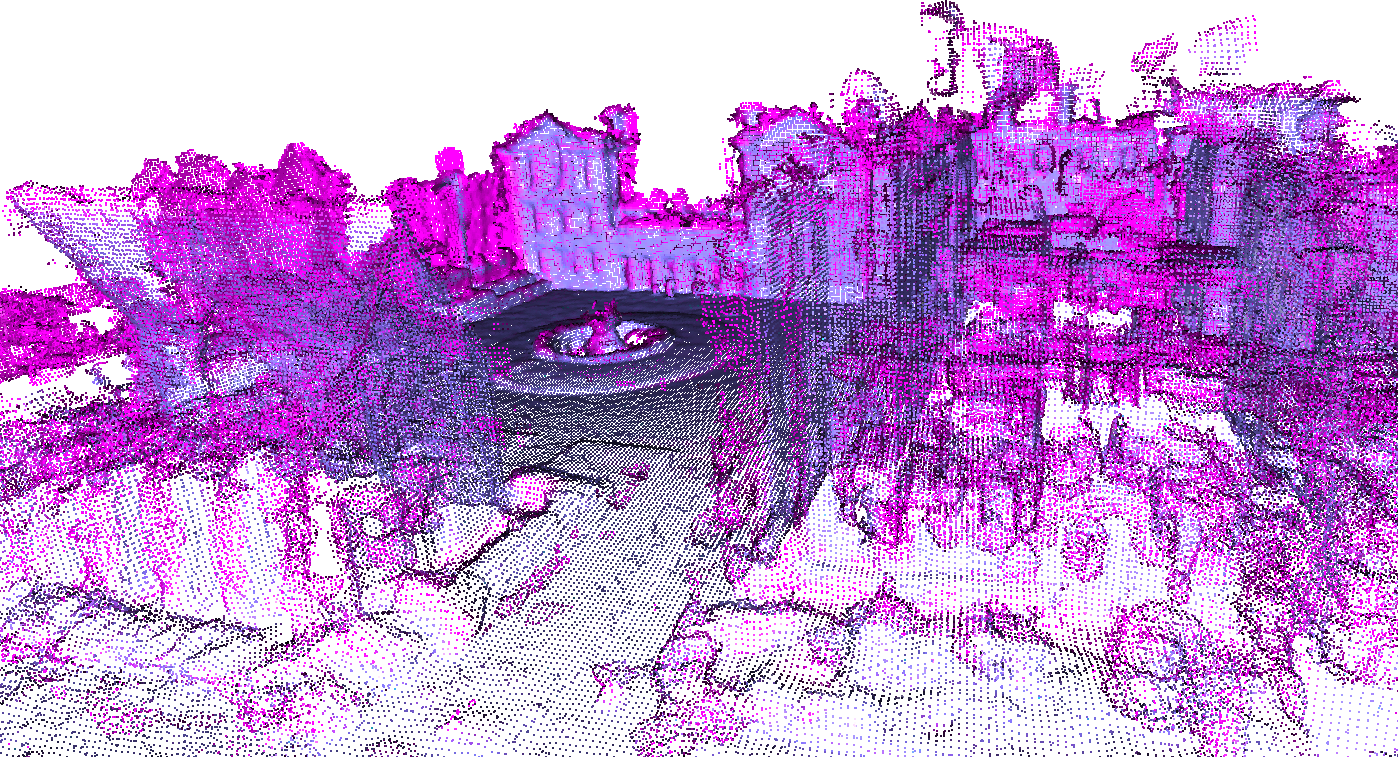}
		\caption{20 cm}
	\end{subfigure}
	\\

	\begin{subfigure}[b]{0.49\linewidth}
		\centering
		\includegraphics[width=\linewidth]{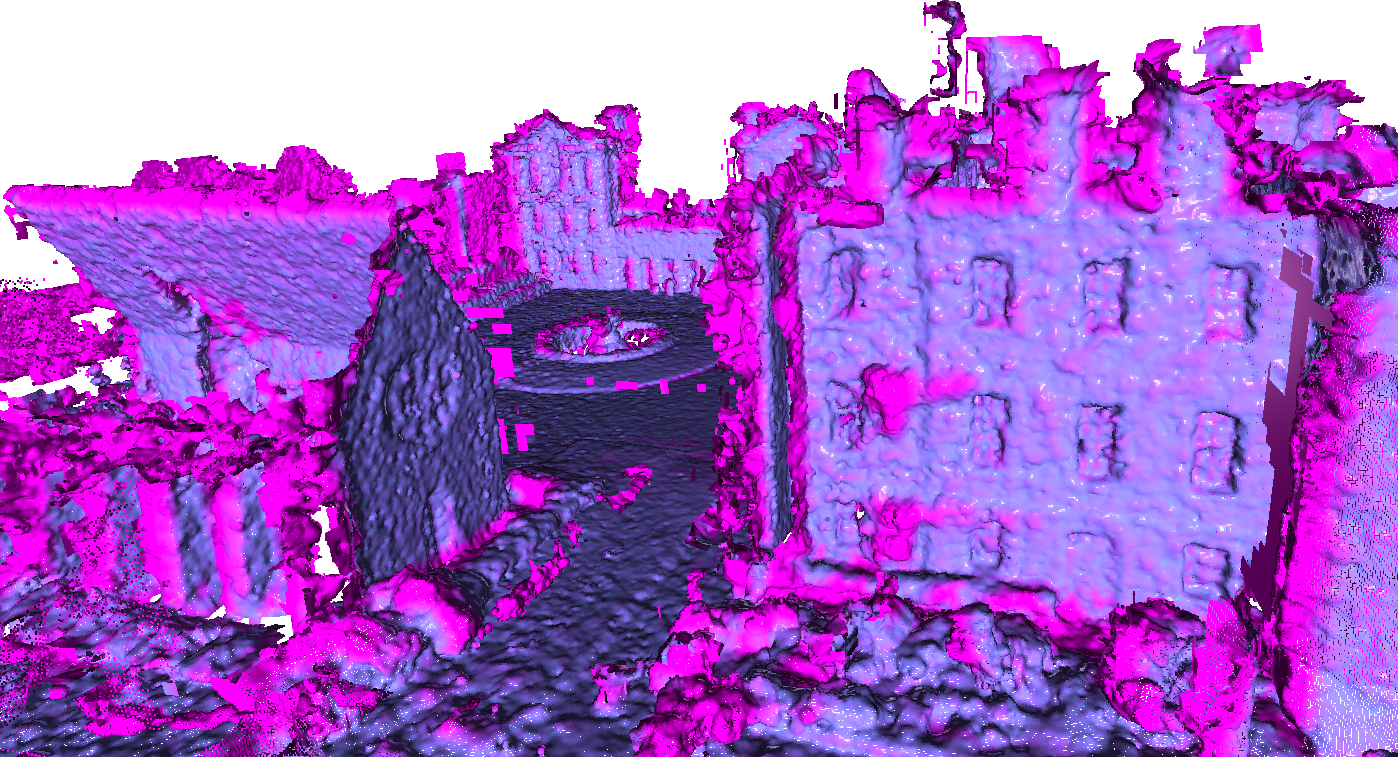}
		\caption{5 cm}
	\end{subfigure}
	\begin{subfigure}[b]{0.49\linewidth}
		\centering
		\includegraphics[width=\linewidth]{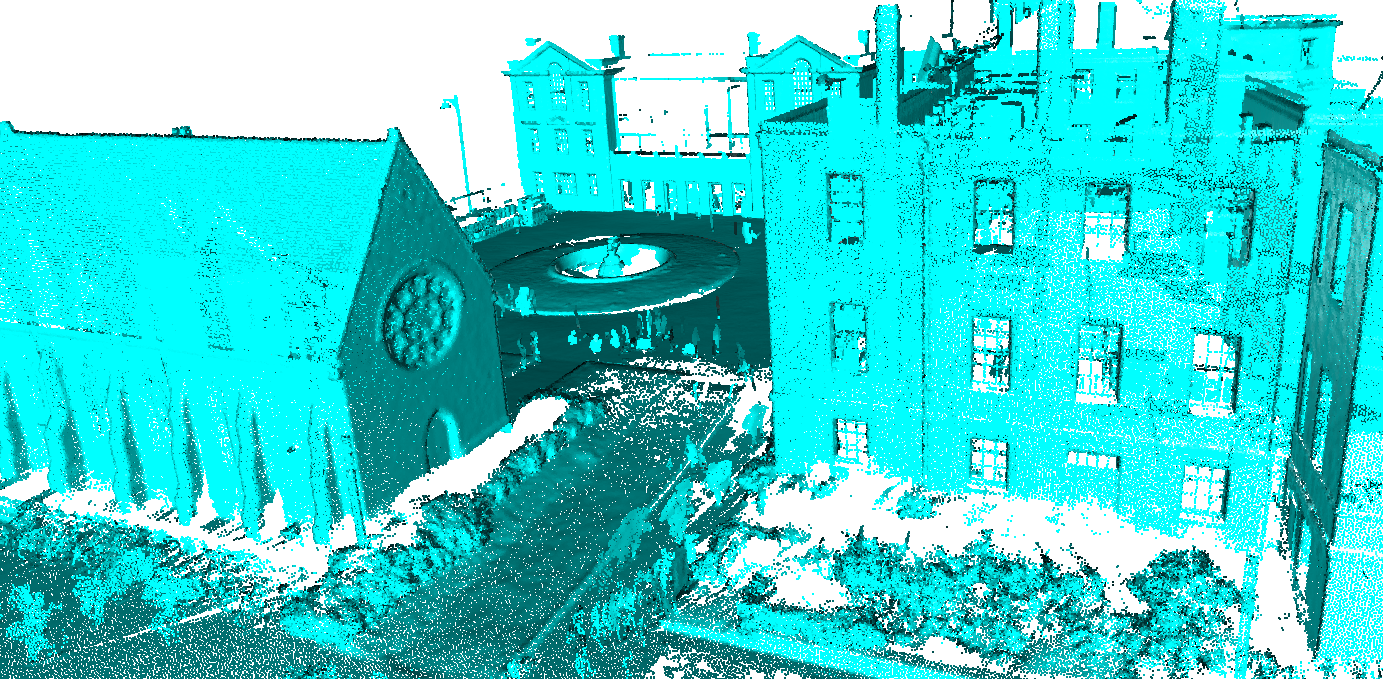}
		\caption{Ground-truth}
	\end{subfigure}
	\caption{Continuous reconstruction of PIN-SLAM with voxel sizes in TABLE \ref{tab:continuous reconstruction}(b).}
	\label{fig:continuous recons pin-slam}
\end{figure}

PIN-SLAM can also extract meshes with varying marching cube resolutions on the SDF map for continuous reconstruction. However, as shown in TABLE \ref{tab:continuous reconstruction}(b), increasing the {voxel grid resolution used to discretize the signed distance field for finer mesh extraction} leads to reduced map accuracy. This suggests that 
more points are being sampled from invalid regions by PIN-SLAM, as illustrated in Fig.~\ref{fig:continuous recons pin-slam} when comparing with Fig.\ref{fig:continuous reconstruction}.

\subsection{ Evaluation on Mapping with Different CURL Degrees}
\begin{figure}[t]
	\centering
	\includegraphics[width=1\linewidth]{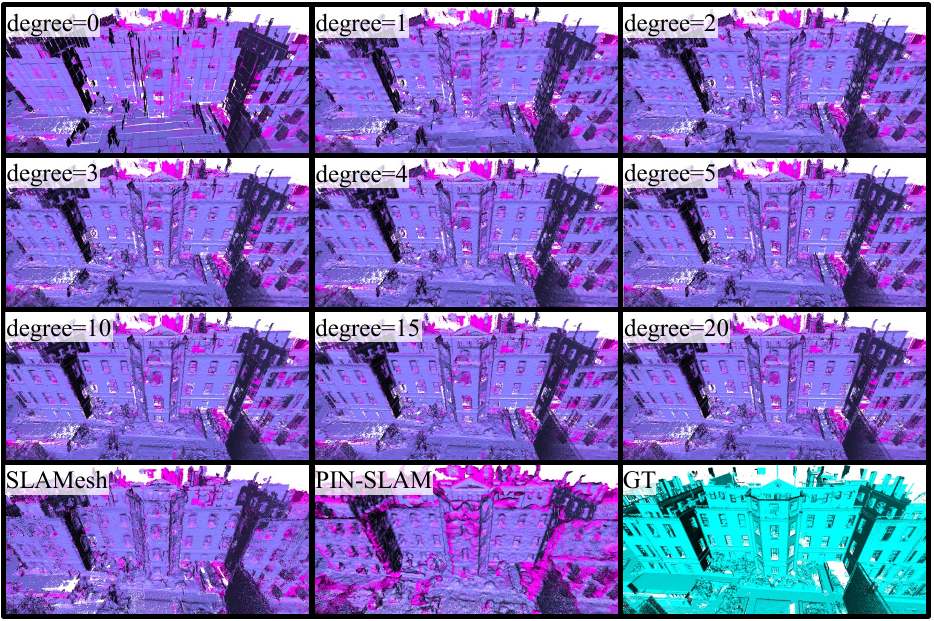}
	\caption{Reconstruction at different CURL degrees. SLAMesh, PIN-SLAM and Ground-truth results are shown in the last row.}
	\label{fig:diff degree}
\end{figure}
\begin{table}[]
	\caption{Map evaluation on CURL degrees (\texttt{math\_e}).}
	\label{tab:different degree recons}
	\centering
	\resizebox{\linewidth}{!}{
		\begin{tabular}{l|ccccccccc}
			\hline
			\multicolumn{1}{l|}{\multirow{2}{*}{Metric}} & \multicolumn{9}{c}{CURL Degree}                                                                                                                                                                                                                                                                                                                                                                                                                                                                                                                                                                                                                                                                                                                                                                                                                                                                                       \\ \cline{2-10}
			                                             & \multicolumn{1}{c|}{0}                                                                             & \multicolumn{1}{c|}{1}                                                                             & \multicolumn{1}{c|}{2}                                                                             & \multicolumn{1}{c|}{3}                                                                             & \multicolumn{1}{c|}{4}                                                                             & \multicolumn{1}{c|}{5}                                                                             & \multicolumn{1}{c|}{10}                                                                            & \multicolumn{1}{c|}{15}                                                                            & 20                                                                            \\ \hline
			M. Acc   $\downarrow$                        & \multicolumn{1}{c|}{\cellcolor[rgb]{0.760000000000000	0.490000000000000	0.580000000000000}{16.82}} & \multicolumn{1}{c|}{\cellcolor[rgb]{0.736250000000000	0.515000000000000	0.545000000000000}{10.33}} & \multicolumn{1}{c|}{\cellcolor[rgb]{0.712500000000000	0.540000000000000	0.510000000000000}{9.95}}  & \multicolumn{1}{c|}{\cellcolor[rgb]{0.688750000000000	0.565000000000000	0.475000000000000}{9.75} } & \multicolumn{1}{c|}{\cellcolor[rgb]{0.665000000000000	0.590000000000000	0.440000000000000}{9.61} } & \multicolumn{1}{c|}{\cellcolor[rgb]{0.641250000000000	0.615000000000000	0.405000000000000}{9.53} } & \multicolumn{1}{c|}{\cellcolor[rgb]{0.570000000000000	0.690000000000000	0.300000000000000}{9.38} } & \multicolumn{1}{c|}{\cellcolor[rgb]{0.593750000000000	0.665000000000000	0.335000000000000}{9.39} } & \cellcolor[rgb]{0.617500000000000	0.640000000000000	0.370000000000000}{9.46 } \\ \hline
			M. Comp  $\downarrow$                        & \multicolumn{1}{c|}{\cellcolor[rgb]{0.760000000000000	0.490000000000000	0.580000000000000}{30.99}} & \multicolumn{1}{c|}{\cellcolor[rgb]{0.736250000000000	0.515000000000000	0.545000000000000}{20.26}} & \multicolumn{1}{c|}{\cellcolor[rgb]{0.712500000000000	0.540000000000000	0.510000000000000}{19.65}} & \multicolumn{1}{c|}{\cellcolor[rgb]{0.688750000000000	0.565000000000000	0.475000000000000}{19.28}} & \multicolumn{1}{c|}{\cellcolor[rgb]{0.665000000000000	0.590000000000000	0.440000000000000}{19.05}} & \multicolumn{1}{c|}{\cellcolor[rgb]{0.641250000000000	0.615000000000000	0.405000000000000}{18.88}} & \multicolumn{1}{c|}{\cellcolor[rgb]{0.617500000000000	0.640000000000000	0.370000000000000}{18.68}} & \multicolumn{1}{c|}{\cellcolor[rgb]{0.570000000000000	0.690000000000000	0.300000000000000}{18.60}} & \cellcolor[rgb]{0.593750000000000	0.665000000000000	0.335000000000000}{18.65} \\ \hline
			C-L1     $\downarrow$                        & \multicolumn{1}{c|}{\cellcolor[rgb]{0.760000000000000	0.490000000000000	0.580000000000000}{82.38}} & \multicolumn{1}{c|}{\cellcolor[rgb]{0.736250000000000	0.515000000000000	0.545000000000000}{15.29}} & \multicolumn{1}{c|}{\cellcolor[rgb]{0.712500000000000	0.540000000000000	0.510000000000000}{14.80}} & \multicolumn{1}{c|}{\cellcolor[rgb]{0.688750000000000	0.565000000000000	0.475000000000000}{14.52}} & \multicolumn{1}{c|}{\cellcolor[rgb]{0.665000000000000	0.590000000000000	0.440000000000000}{14.33}} & \multicolumn{1}{c|}{\cellcolor[rgb]{0.641250000000000	0.615000000000000	0.405000000000000}{14.21}} & \multicolumn{1}{c|}{\cellcolor[rgb]{0.593750000000000	0.665000000000000	0.335000000000000}{14.03}} & \multicolumn{1}{c|}{\cellcolor[rgb]{0.570000000000000	0.690000000000000	0.300000000000000}{14.00}} & \cellcolor[rgb]{0.617500000000000	0.640000000000000	0.370000000000000}{14.05} \\ \hline
			F-Score  $\uparrow$                          & \multicolumn{1}{c|}{\cellcolor[rgb]{0.760000000000000	0.490000000000000	0.580000000000000}{76.88}} & \multicolumn{1}{c|}{\cellcolor[rgb]{0.736250000000000	0.515000000000000	0.545000000000000}{81.17}} & \multicolumn{1}{c|}{\cellcolor[rgb]{0.712500000000000	0.540000000000000	0.510000000000000}{82.46}} & \multicolumn{1}{c|}{\cellcolor[rgb]{0.688750000000000	0.565000000000000	0.475000000000000}{83.16}} & \multicolumn{1}{c|}{\cellcolor[rgb]{0.665000000000000	0.590000000000000	0.440000000000000}{83.60}} & \multicolumn{1}{c|}{\cellcolor[rgb]{0.641250000000000	0.615000000000000	0.405000000000000}{83.89}} & \multicolumn{1}{c|}{\cellcolor[rgb]{0.617500000000000	0.640000000000000	0.370000000000000}{84.38}} & \multicolumn{1}{c|}{\cellcolor[rgb]{0.570000000000000	0.690000000000000	0.300000000000000}{84.47}} & \cellcolor[rgb]{0.593750000000000	0.665000000000000	0.335000000000000}{84.40} \\ \hline
		\end{tabular}}
	\begin{flushleft}
		{\footnotesize * The greener, the better result.}
	\end{flushleft}
\end{table}
Our map accuracy depends on the degree of spherical harmonics coefficients used. The higher the degree, the more geometric variation can be captured. Therefore, we evaluate the relationship between map accuracy and the degree of the CURL's spherical harmonics.

As shown in Fig. \ref{fig:diff degree}, at degree 0, all patches are represented as flat planes. As the degree increases, more details are recovered. TABLE \ref{tab:different degree recons} supports the same finding: with increasing CURL degrees, map accuracy gradually improves. However, the improvement becomes marginal beyond degree 5, indicating that degree 5 captures most of the geometric details for scenarios with similar structural complexity as the sequence \texttt{math\_e}.

\subsection{Evaluation on Map Size}
\begin{table}[]
	\caption{Map size evaluation (Megabyte - MB)}
	\label{tab:map size}
	\centering
	\begin{tabular}{ll|c|c|c|c}
		\hline
		\multicolumn{2}{c|}{Datasets}                                                  & HBA            & SLAMesh & PIN-SLAM & Ours                               \\ \hline
		\multicolumn{1}{l|}{\multirow{6}{*}{\rotatebox[origin=c]{90}{Newer College}}}  & quad\_e        & 198.8   & 29.5     & \underline{10.8} & $\mathbf{3.5 }$ \\ \cline{2-6}
		\multicolumn{1}{l|}{}                                                          & math\_e        & 274.7   & 62.9     & \underline{19.4} & $\mathbf{8.6 }$ \\ \cline{2-6}
		\multicolumn{1}{l|}{}                                                          & ug\_e          & 71.3    & 13.8     & \underline{3.3 } & $\mathbf{2.0 }$ \\ \cline{2-6}
		\multicolumn{1}{l|}{}                                                          & cloister       & 227.1   & 41.9     & \underline{9.1 } & $\mathbf{6.3 }$ \\ \cline{2-6}
		\multicolumn{1}{l|}{}                                                          & stairs         & 18.1    & N/A      & \underline{4.0 } & $\mathbf{1.9 }$ \\ \cline{2-6}
		\multicolumn{1}{l|}{}                                                          & parkland       & 807.5   & 147.8    & \underline{43.4} & $\mathbf{21.8}$ \\ \hline
		\multicolumn{1}{l|}{\multirow{8}{*}{\rotatebox[origin=c]{90}{FusionPortable}}} & canteen\_night & 192.6   & 53.5     & \underline{11.7} & $\mathbf{4.3 }$ \\ \cline{2-6}
		\multicolumn{1}{l|}{}                                                          & garden\_night  & 208.2   & 50.0     & \underline{9.5 } & $\mathbf{3.3 }$ \\ \cline{2-6}
		\multicolumn{1}{l|}{}                                                          & MCR\_fast      & 33.6    & 3.6      & \underline{1.5 } & $\mathbf{0.47}$ \\ \cline{2-6}
		\multicolumn{1}{l|}{}                                                          & MCR\_normal    & 28.7    & 3.0      & \underline{1.3 } & $\mathbf{0.44}$ \\ \cline{2-6}
		\multicolumn{1}{l|}{}                                                          & MCR\_slow      & 27.8    & 3.0      & \underline{1.2 } & $\mathbf{0.42}$ \\ \cline{2-6}
		\multicolumn{1}{l|}{}                                                          & corridor\_day  & 170.3   & 49.3     & \underline{9.4 } & $\mathbf{6.9 }$ \\ \cline{2-6}
		\multicolumn{1}{l|}{}                                                          & escalator\_day & 298.0   & 65.7     & \underline{12.9} & $\mathbf{10.0}$ \\ \cline{2-6}
		\multicolumn{1}{l|}{}                                                          & building\_day  & 591.7   & 112.4    & \underline{31.4} & $\mathbf{17.2}$ \\ \hline
	\end{tabular}
	\begin{flushleft}
		{\footnotesize The best results are highlighted in \textbf{bold}, while the second-best results are \underline{underlined}.}
	\end{flushleft}
\end{table}

\begin{figure}[t]
	\centering
	\includegraphics[width=1\linewidth]{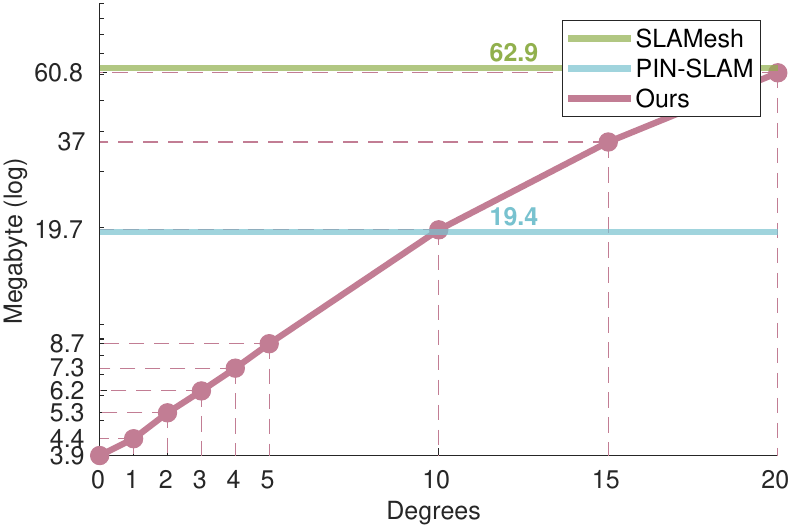}
	\caption{Map size changes with degrees (\texttt{math\_e}).}
	\label{fig:size_vs_degrees}
\end{figure}

For each patch, we only need to store a boolean value to indicate whether it is a ground or non-ground patch, which helps to determine the degree it utilizes. Then, the corresponding spherical harmonics coefficients are stored, consuming $(L+1)^2$ double-precision numbers. Additionally, a transformation matrix of size $3\times 4$ floats is stored to represent the information of $\mathbf{T}_{\mathtt{W}\mathtt{M}_k}$ for the $k$-th patch. Finally, a binary mask of size $\omega^2$ bits is required. For a map with $N$ patches, the map size in total can be formulated as follows:
\begin{equation}
	\small
	\begin{aligned}
		\text{map\_size} = & N \times \Big( \text{sizeof}(\text{bool})
		+ (L+1)^2 \times \text{sizeof}(\text{double})                  \\
		                   & + 12 \times \text{sizeof}(\text{float})
		+ \frac{\omega^2}{8} \Big)\ \text{bytes}
	\end{aligned}
\end{equation}
Therefore, by saving the data in the aforementioned format as a binary file, we can achieve an ultra-compact map. This saved map also supports continuous reconstruction with varying resolutions.

From TABLE \ref{tab:map size}, we observe that our map size is approximately 2.8\% of HBA's dense point cloud map, 13.4\% of SLAMesh's mesh map  and around 51.6\% of PIN-SLAM's neural map on average.

Our map size heavily depends on the degree of spherical harmonics used, as the number of coefficients equals $(L+1)^2$. Fig. \ref{fig:size_vs_degrees} illustrates how our map size changes with different degrees. It shows that PIN-SLAM's map size is similar to ours when using a degree of 10, while SLAMesh's map size is comparable to our map size with a degree of 20.

\subsection{Evaluation on Estimated Trajectory}

\begin{table}[h!]
	\centering
	\caption{Odometry evaluation}
	\label{tab:RE_trajectory}
	\centering
	\begin{adjustbox}{width=\linewidth}
		\renewcommand{\arraystretch}{1.1}
		{\begin{tabular}{ll|c|c|c}
				\hline
				\multicolumn{2}{c|}{Datasets}                                                  & SLAMesh        & PIN-SLAM                             & Ours                                                                      \\ \hline
				\multicolumn{1}{l|}{\multirow{5}{*}{\rotatebox[origin=c]{90}{Newer College}}}  & quad\_e        & 0.427/1.539                          & \textbf{0.277}/\textbf{1.163}       & \underline{0.394}/\underline{1.396} \\ \cline{2-5}
				\multicolumn{1}{l|}{}                                                          & math\_e        & \textbf{0.252}/0.982                 & \underline{0.265}/\textbf{0.512}    & 0.272/\underline{0.702}             \\ \cline{2-5}
				\multicolumn{1}{l|}{}                                                          & ug\_e          & 1.000/5.532                          & \textbf{0.383}/\textbf{2.020}       & \underline{0.881}/\underline{4.630} \\ \cline{2-5}
				\multicolumn{1}{l|}{}                                                          & cloister*      & 0.992/2.610                          & \textbf{0.423}/\textbf{1.042}       & \underline{0.442}/\underline{1.706} \\ \cline{2-5}
				\multicolumn{1}{l|}{}                                                          & parkland*      & \underline{0.784}/\underline{0.926}  & 1.735/1.954                         & \textbf{0.637}/\textbf{0.814}       \\ \hline
				\multicolumn{1}{c|}{\multirow{7}{*}{\rotatebox[origin=c]{90}{FusionPortable}}} & canteen\_night & \underline{0.115}/\underline{0.650}  & 0.154/\textbf{0.647}                & \textbf{0.114}/0.666                \\ \cline{2-5}
				\multicolumn{1}{c|}{}                                                          & garden\_night  & 0.254/1.384                          & \underline{0.149}/\underline{0.787} & \textbf{0.111}/\textbf{0.638}       \\ \cline{2-5}
				\multicolumn{1}{l|}{}                                                          & MCR\_fast      & 0.246/4.647                          & \textbf{0.176}/\underline{3.631}    & \underline{0.193}/\textbf{3.186}    \\ \cline{2-5}
				\multicolumn{1}{l|}{}                                                          & MCR\_normal    & \underline{0.295}/\underline{80.056} & \textbf{0.290}/\textbf{79.751}      & 0.337/80.113                        \\ \cline{2-5}
				\multicolumn{1}{l|}{}                                                          & corridor\_day  & 2.019/4.303                          & \textbf{0.398}/\textbf{0.879}       & \underline{1.087}/\underline{1.620} \\ \cline{2-5}
				\multicolumn{1}{l|}{}                                                          & escalator\_day & \underline{0.233}/\underline{1.911}  & \textbf{0.180}/\textbf{1.077}       & 0.243/1.925                         \\ \cline{2-5}
				\multicolumn{1}{l|}{}                                                          & building\_day  & \textbf{0.329}/\underline{0.607}     & 0.438/0.733                         & \underline{0.337}/\textbf{0.601}    \\ \hline
			\end{tabular}}
	\end{adjustbox}
	\begin{flushleft}
		{\footnotesize Relative errors averaged over trajectories of 100 to 800 m length: relative translational error in \% / relative rotational error in degrees per 100 m.\\
			The best results are highlighted in \textbf{bold}, while the second-best results are \underline{underlined}.\\
			* represents the sequences that using all the available associated patches for pose estimation in our method.}
	\end{flushleft}
\end{table}

\begin{table}[]
	\caption{Trajectory evaluation (ATE in $^{\circ}$/meter)}
	\label{tab:ATE_trajectory}
	\centering
	\begin{adjustbox}{width=\linewidth}
		\renewcommand{\arraystretch}{1.1}
		{ \begin{tabular}{ll|c|c|c|c}
				\hline
				\multicolumn{2}{c|}{Datasets}                                                  & HBA            & SLAMesh                             & PIN-SLAM                         & Ours                                                                       \\ \hline
				\multicolumn{1}{l|}{\multirow{6}{*}{\rotatebox[origin=c]{90}{Newer College}}}  & quad\_e        & \textbf{0.978}/0.087                & 1.337/\textbf{0.084}             & \underline{0.986}/0.104              & 1.225/\underline{0.086}             \\ \cline{2-6}
				\multicolumn{1}{l|}{}                                                          & math\_e        & \underline{0.440}/\textbf{0.073}    & 1.010/0.116                      & \textbf{0.420}/\underline{0.080}     & 0.708/\textbf{0.073}                \\ \cline{2-6}
				\multicolumn{1}{l|}{}                                                          & ug\_e          & \underline{1.628}/\textbf{0.074}    & 4.120/0.262                      & \textbf{1.567}/\textbf{0.074}        & 5.587/\underline{0.180}             \\ \cline{2-6}
				\multicolumn{1}{l|}{}                                                          & cloister*      & \underline{1.277}/\underline{0.137} & 3.140/0.242                      & \textbf{1.249}/\textbf{0.134}        & 3.200/0.212                         \\ \cline{2-6}
				\multicolumn{1}{l|}{}                                                          & stairs*        & 10.422/1.078                        & N/A                              & \textbf{1.392}/\textbf{0.075}        & \underline{5.320}/\underline{0.169} \\ \cline{2-6}
				\multicolumn{1}{l|}{}                                                          & parkland*      & 3.216/\underline{0.288}             & \underline{3.139}/0.486          & \textbf{1.261}/0.345                 & 3.420/\textbf{0.271}                \\ \hline
				\multicolumn{1}{c|}{\multirow{8}{*}{\rotatebox[origin=c]{90}{FusionPortable}}} & canteen\_night & \textbf{0.685}/\textbf{0.078}       & \underline{0.693}/\textbf{0.078} & 0.700/0.109                          & 0.702/\underline{0.081}             \\ \cline{2-6}
				\multicolumn{1}{c|}{}                                                          & garden\_night  & \underline{0.660}/\textbf{0.038}    & 1.412/0.077                      & 0.758/0.091                          & \textbf{0.640}/\underline{0.040}    \\ \cline{2-6}
				\multicolumn{1}{l|}{}                                                          & MCR\_fast      & \textbf{14.087}/\textbf{0.168}      & 14.539/0.195                     & \underline{14.120}/\underline{0.173} & 14.551/{0.184}                      \\ \cline{2-6}
				\multicolumn{1}{l|}{}                                                          & MCR\_normal    & \underline{28.668}/0.116            & 28.773/\underline{0.114}         & \textbf{28.615}/\textbf{0.107}       & 28.737/0.115                        \\ \cline{2-6}
				\multicolumn{1}{l|}{}                                                          & MCR\_slow      & 17.132/0.088                        & \textbf{16.962}/\textbf{0.061}   & \underline{17.122}/\underline{0.077} & 17.262/0.096                        \\ \cline{2-6}
				\multicolumn{1}{l|}{}                                                          & corridor\_day  & \textbf{1.086}/\textbf{0.125}       & 8.767/3.000                      & \underline{1.232}/\underline{0.189}  & 2.570/0.304                         \\ \cline{2-6}
				\multicolumn{1}{l|}{}                                                          & escalator\_day & \textbf{1.453}/\textbf{0.181}       & 2.733/\underline{0.186}          & \underline{1.461}/\underline{0.186}  & 4.875/0.187                         \\ \cline{2-6}
				\multicolumn{1}{l|}{}                                                          & building\_day  & 1.473/0.154                         & \textbf{1.284}/\textbf{0.131}    & 1.568/0.184                          & \underline{1.383}/\underline{0.141} \\ \hline
			\end{tabular}}
	\end{adjustbox}
	\begin{flushleft}
		{\footnotesize The best results are highlighted in \textbf{bold}, while the second-best results are \underline{underlined}. N/A indicates a failure.\\
			* represents the sequences that using all the available associated patches for pose estimation in our method.}
	\end{flushleft}
\end{table}

\subsubsection{{LiDAR Odometry Evaluation}}
{TABLE~\ref{tab:RE_trajectory} presents the pure odometry accuracy, obtained by disabling the loop closure modules in both PIN-SLAM and our method. Across most sequences, our method consistently ranks second in performance. The HBA method is excluded from this comparison, as it does not include an odometry module. Additionally, the \texttt{stairs} and \texttt{MCR\_slow} sequences are omitted as they are too short to provide meaningful results.}

\subsubsection{{LiDAR SLAM Trajectory Evaluation}}
TABLE \ref{tab:ATE_trajectory} demonstrates that the ATE trajectory accuracy of our proposed method consistently ranks in the top two positions in over half of all the sequences, achieving localization accuracy comparable to that of HBA, SLAMesh and PIN-SLAM. {Notably, the ATE results reported here are from the same runs used in the map reconstruction evaluation, ensuring consistency across trajectory and mapping performance comparisons.}
{Although the HBA method generally achieves the best trajectory accuracy, it performs the worst in the map accuracy. This observation tentatively suggests that, after a certain level of pose precision is reached, additional factors, such as the underlying map representation, may exert a greater influence on mapping quality. A thorough exploration of this hypothesis lies beyond the scope of the present work, but it nonetheless suggests that the map representation is needing for more attention, and partially motivates this work.}

SLAMesh, while performing well in several sequences (e.g., \texttt{canteen\_night}, \texttt{building\_day}), shows inconsistent performance, particularly due to the lack of loop closure, as evidenced in the \texttt{corridor\_day} sequence.
PIN-SLAM achieves stable and robust results across all sequences, demonstrating consistent performance.
Notably, our algorithm requires 
all associated patches for localization in scenarios with narrow spaces and limited FoV, like \texttt{cloister} and \texttt{stairs} sequences.

\subsection{{Effect of Local Bundle Adjustment}}

{Additionally, to further evaluate the effect of BA, we compare ATE results on the same runs by disabling and enabling local BA for sequences where loop closures are detected. The quantitative results, presented in TABLE \ref{tab:local_ba_ate}, highlight the impact of local BA on trajectory accuracy. As shown in TABLE \ref{tab:local_ba_ate}, enabling local BA consistently improves or maintains trajectory accuracy across all evaluated sequences. The improvement is especially notable in more challenging scenarios such as \texttt{stairs}, where the ATE is significantly reduced from 0.509 m to 0.372 m. These results demonstrate the effectiveness of local BA in refining trajectory estimates after loop closure, particularly in complex environments with narrow spaces or limited field of view.}

\begin{table}
	\centering
	\caption{ATE (m) with/without Local BA}
	\label{tab:local_ba_ate}
	{\begin{tabular}{l|c|c}
			\hline
			\multicolumn{1}{l|}{Sequences} & Pose-Graph Only & With Local BA  \\ \hline
			math\_e                        & 0.072           & \textbf{0.067} \\ \hline
			ug\_e                          & 0.078           & \textbf{0.077} \\ \hline
			cloister                       & 0.205           & \textbf{0.204} \\ \hline
			stairs                         & 0.509           & \textbf{0.372} \\ \hline
			parkland                       & 0.240           & \textbf{0.234} \\ \hline
			corridor\_day                  & 0.168           & \textbf{0.167} \\ \hline
			escalator\_day                 & 0.195           & \textbf{0.193} \\ \hline
			building\_day                  & 0.121           & \textbf{0.120} \\ \hline
		\end{tabular}}
	\begin{flushleft}
		{\hspace{4em}\footnotesize The best results are highlighted in \textbf{bold}.}
	\end{flushleft}
\end{table}

\subsection{Evaluation on Runtime}

\begin{figure}[t]
	\centering
	\begin{subfigure}[b]{\linewidth}
		\centering
		\includegraphics[width=\linewidth]{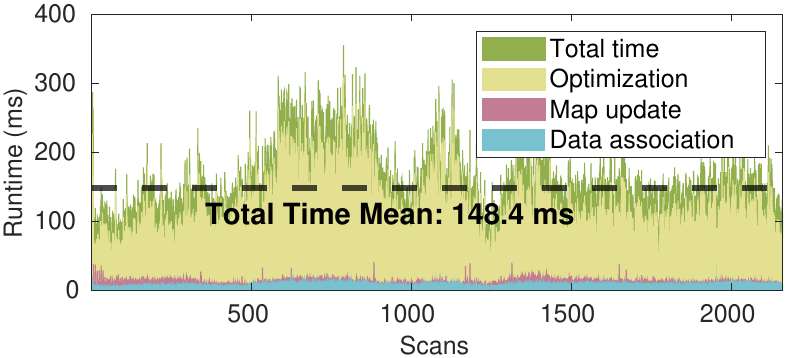}
		\caption{Full version ($0.322\%$ and $0.818^\circ$).} \label{fig:runtime_eval_full}
	\end{subfigure}
	\begin{subfigure}[b]{\linewidth}
		\centering
		\includegraphics[width=\linewidth]{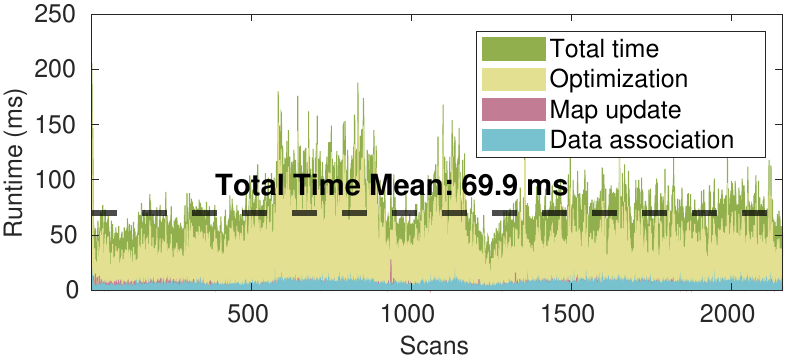}
		\caption{Light version ($0.330\%$ and $0.797^\circ$). }\label{fig:runtime_eval}
	\end{subfigure}
	\caption{Runtime evaluation of CURL-SLAM on \texttt{math\_e}, with relative translation and rotation errors in brackets.}
\end{figure}

Fig. \ref{fig:runtime_eval_full} shows our runtime performance using the default parameters. All modules run on a single thread, except for the map update, which uses 16 threads, achieving 6.74 Hz on a CPU. By setting $\omega=16$, $\beta_{ng}=15$, and $\beta_{g}=10$ to reduce the patch resolution and the limit on associated pairs, 
achieving real-time performance of 14.3Hz, as shown in Fig. \ref{fig:runtime_eval}, while maintaining similar odometry performance. 
This test was conducted on a 128-channel LiDAR; fewer channels would further reduce processing time. 

Both PIN-SLAM and our proposed method are capable of achieving globally consistent map reconstruction at multiple resolutions and encoding the map into a compact representation. However, PIN-SLAM is limited to real-time performance only when utilizing 
certain GPU architectures. Additionally, extracting mesh maps for large-scale regions with PIN-SLAM demands substantial amounts of RAM memory, making it impractical for embedded systems. Although PIN-SLAM can be operated on a CPU, it requires processing times exceeding two seconds per scan, which restricts its applicability in real-time and resource-constrained environments. In contrast, our method offers more flexibility and efficiency, particularly in scenarios where hardware resources are limited.

\section{Conclusions}

This work proposes a novel LiDAR mapping system, CURL-SLAM, using the CURL representation with the full pipeline of odometry, local bundle adjustment and loop closure. It is capable of continuously generating 3D reconstructions at a desired resolution. To the best of our knowledge, our method achieves the most efficient map size in dense online LiDAR SLAM systems, with the capability of continuous dense 3D reconstruction.
Our experiments demonstrate that the proposed CURL-SLAM method achieves the state-of-the-art mapping performance and comparable localization accuracy. Finally, our system can operate in real-time on a single Intel i7-10875H CPU.

In future work, we will investigate the integration of Inertial Measurement Unit with CURL representation, aiming to improve the robustness of pose estimation in environments with limited field of view. 



\addtolength{\textheight}{0cm}

\section*{Appendix A}
\label{sec:appendix a}
This appendix provides the terms in \eqref{spherical_harmonics}, \eqref{eq:Jac}, \eqref{eq:J_I_q}, and \eqref{eq:J_I_j} that were not given in the text.
\begin{equation}
	N_m(\phi)        =\left\{
	\begin{matrix}
		\sqrt{2}cos(m\phi),   & m>0 \\
		1,                    & m=0 \\
		\sqrt{2}sin(|m|\phi), & m<0 \\
	\end{matrix}
	\right.
\end{equation}

\begin{equation}
	\mathbf{J}^ {\mathsf{P}_z}=[\begin{matrix}
				0 & 0 & 1
			\end{matrix}]
	\label{eq:Z}
\end{equation}
The numerical jacobian for $\mathbf{I}^k$ is
\begin{equation}
	\mathbf{J}^ {\mathbf{I}^k} = \left[
		\begin{matrix}
			\mathbf{J}_x^{\mathbf{I}^k} & \mathbf{J}_y^{\mathbf{I}^k}
		\end{matrix}
		\right]
	\label{eq:numerical I}
\end{equation}
where
$$\mathbf{J}_x^{\mathbf{I}^k} \approx
	\frac{\mathbf{I}^k([x+\delta ,y]^T)-\mathbf{I}^k([x-\delta ,y]^T)}{2\delta}$$
$$\mathbf{J}_y^{\mathbf{I}^k} \approx
	\frac{\mathbf{I}^k([x ,y+\delta]^T)-\mathbf{I}^k([x ,y-\delta]^T)}{2\delta}$$
and $\delta$ is a small constant.

\begin{equation}
	\mathbf{J}^ {\mathsf{P}_{xy}}=\left[
		\begin{matrix}
			1 & 0 & 0 \\
			0 & 1 & 0
		\end{matrix}
		\right]
	\label{eq:XY}
\end{equation}



\begin{equation}
	\mathbf{J}^{\mathsf{P}_x} = \left[
		1\quad 0
		\right] \qquad
	\mathbf{J}^{\mathsf{P}_y} = \left[
		0\quad 1
		\right]
\end{equation}

\begin{equation}
	\mathbf{J}^{\phi} = \frac{2\pi \eta}{s}\qquad
	\mathbf{J}^{\theta} = \frac{\pi \eta}{s}
\end{equation}

\section*{Appendix B}
\label{sec:appendix b}

\textbf{Lemma 1.} For a transformation matrix $\mathbf{T} = \left[
		\begin{matrix}
			\mathbf{R} & \mathbf{t} \\
			\mathbf{0} & 1
		\end{matrix}
		\right]
	\in SE(3) \subset \mathbb{R}^{4\times 4}$, which acts on a point $\mathbf{p} \in \mathbb{R}^{3 \times 1}$, such that $\mathbf{T} \cdot \mathbf{p} = \mathbf{R}\mathbf{q} + \mathbf{t}$. The Jacobian matrix $\mathbf{J}_{\mathbf{T}}^{\mathbf{T \cdot q}}$ is given by:
\begin{equation}
	\mathbf{J}_{\mathbf{T}}^{\mathbf{T \cdot q}} = \left[ \mathbf{R} \quad -\mathbf{R} [\mathbf{p}]_{\times} \right]
	\label{eq:lemma1}
\end{equation}
where $[\cdot]_{\times}$ is the skew-symmetric matrix of the vector.

\textbf{Lemma 2.} For the transformation matrices $\mathbf{T}_1 = \left[
		\begin{matrix}
			\mathbf{R}_1 & \mathbf{t}_1 \\
			\mathbf{0}   & 1
		\end{matrix}
		\right]
	\in SE(3) \subset \mathbb{R}^{4\times 4}$ and $\mathbf{T}_2 = \left[
		\begin{matrix}
			\mathbf{R}_2 & \mathbf{t}_2 \\
			\mathbf{0}   & 1
		\end{matrix}
		\right]
	\in SE(3) \subset \mathbb{R}^{4\times 4}$, which act on a point $\mathbf{p} \in \mathbb{R}^{3\times 1}$, the Jacobian matrix $\mathbf{J}_{\mathbf{T}_2}^{\mathbf{T}_1\mathbf{T}_2^{-1}\cdot \mathbf{p}}$ is given by:

\begin{equation}
	\mathbf{J}_{\mathbf{T}_2}^{\mathbf{T}_1\mathbf{T}_2^{-1}\cdot \mathbf{p}} =
	\left[
	-\mathbf{R}_1 \quad \mathbf{R}_1[\mathbf{T}_2^{-1} \cdot \mathbf{p}]_{\times}
	\right]
	\label{eq:lemma2}
\end{equation}

\textbf{Lemma 3.} For function $Y_{l,m}(\theta, \phi)$, its Jacobian matrix $\mathbf{J}^{y_{l,m}}$ is

\begin{equation}
	\begin{aligned}
		\mathbf{J}^{Y_{l,m}} = \left[
		\mathbf{J}_{\phi}^{Y_{l,m}}\quad
		\mathbf{J}_{\theta}^{Y_{l,m}}
		\right] \in \mathbb{R}^{1\times 2}
	\end{aligned}
	\label{eq:J_Y}
\end{equation}

\begin{equation}
	\begin{aligned}
		\mathbf{J}_{\phi}^{Y_{l,m}} =
		\sqrt{\frac{2l+1}{4\pi}\frac{(l-|m|)!}{(l+|m|)!}}P_{l,|m|}(\cos\theta)
		\cdot \\
		\left\{
		\begin{array}{cc}
			-m\sqrt{2}\sin(m\phi), & m>0 \\
			0,                     & m=0 \\
			-m\sqrt{2}\cos(m\phi), & m<0
		\end{array}
		\right.
	\end{aligned}
	\label{eq:J_Y_phi}
\end{equation}

\begin{equation}
	\small
	\begin{aligned}
		\mathbf{J}_{\theta}^{Y_{l,m}} = \left\{
		\begin{array}{ll}
			mcot\theta Y_{l,m}(\theta,\phi)-                                                    \\
			\sqrt{\frac{2l+1}{4\pi}\frac{(l-m)!}{(l+m)!}}N_m(\phi) P_{l,m+1}(\cos\theta), & m>0 \\
			\frac{-\sqrt{l(l+1)}}{N_1(\phi)}Y_{l,1}(\theta,\phi)\ or                            \\
			-\sqrt{\frac{2l+1}{4\pi}}P_{l,1}(cos \theta),                                 & m=0 \\
			-mcot\theta Y_{l,m}(\theta,\phi)-                                                   \\
			\sqrt{\frac{2l+1}{4\pi}\frac{(l+m)!}{(l-m)!}}N_m(\phi)P_{l,1-m}(cos\theta),   & m<0
		\end{array}
		\right.
	\end{aligned}
	\label{eq:J_Y_theta}
\end{equation}

\eqref{eq:J_Y_theta} is derived according to \eqref{eq:Legendre} from \cite{bosch2000computation}.

\begin{equation}
	\label{eq:Legendre}
	sin\theta \frac{dP_{l,m}}{d\theta} = m cos\theta P_{l,m} - sin\theta P_{l,m+1}
\end{equation}

\bibliographystyle{IEEEtran}
\bibliography{refs}

\end{document}